\documentclass[letterpaper]{article} 
\usepackage{aaai25}  
\usepackage{times}  
\usepackage{helvet}  
\usepackage{courier}  
\usepackage[hyphens]{url}  
\usepackage{graphicx} 
\usepackage{natbib}  
\usepackage{caption} 
\usepackage[table]{xcolor}
\usepackage{amsmath}
\usepackage{amssymb}
\usepackage{threeparttable}
\usepackage{multirow} 
\usepackage{colortbl}
\usepackage{graphicx}
\usepackage{algorithm}
\usepackage{algorithmic}
\usepackage{pifont}
\usepackage{booktabs}
\usepackage{array}
\usepackage{adjustbox}
\usepackage{threeparttable}

\newcommand{\update}[1]{{\textcolor{black}{#1}}}

\usepackage{newfloat}
\usepackage{listings}
\DeclareCaptionStyle{ruled}{labelfont=normalfont,labelsep=colon,strut=off} 
\floatstyle{ruled}
\newfloat{listing}{tb}{lst}{}
\floatname{listing}{Listing}
\title{HDT: Hierarchical Discrete Transformer for Multivariate Time Series Forecasting}
\author{
    Shibo Feng\textsuperscript{\rm 1, 2}, 
    Peilin Zhao\textsuperscript{\rm 3}\thanks{Corresponding author}, 
    Liu Liu\textsuperscript{\rm 3}, 
    Pengcheng Wu\textsuperscript{\rm 2}, 
    Zhiqi Shen\textsuperscript{\rm 1}\footnotemark[1]
}
\affiliations{
    \textsuperscript{\rm 1}\normalsize{College of Computing and Data Science, Nanyang Technological University (NTU), Singapore} \\
    \textsuperscript{\rm 2}\normalsize{Webank-NTU Joint Research Institute on Fintech, NTU, Singapore}\\
    \textsuperscript{\rm 3}\normalsize{Tencent AI Lab, Shenzhen, China}\\
    \{shibo001, pengcheng.wu, zqshen\}@ntu.edu.sg, \{leonliuliu, masonzhao\}@tencent.com

}

\usepackage{bibentry}

\begin{document}

\maketitle

\begin{abstract}
Generative models have gained significant attention in multivariate time series forecasting (MTS), particularly due to their ability to generate high-fidelity samples. Forecasting the probability distribution of multivariate time series is a challenging yet practical task. Although some recent attempts have been made to handle this task, two major challenges persist: 1) some existing generative methods underperform in high-dimensional multivariate time series forecasting, which is hard to scale to higher dimensions; 2) the inherent high-dimensional multivariate attributes constrain the forecasting lengths of existing generative models. In this paper, we point out that discrete token representations can model high-dimensional MTS with faster inference time, and forecasting the target with long-term trends of itself can extend the forecasting length with high accuracy. Motivated by this, we propose a vector quantized framework called \textbf{H}ierarchical \textbf{D}iscrete \textbf{T}ransformer (HDT) that models time series into discrete token representations with $\ell_2$ normalization enhanced vector quantized strategy, in which we transform the MTS forecasting into discrete tokens generation. To address the limitations of generative models in long-term forecasting, we propose a hierarchical discrete Transformer. This model captures the discrete long-term trend of the target at the low level and leverages this trend as a condition to generate the discrete representation of the target at the high level that introduces the features of the target itself to extend the forecasting length in high-dimensional MTS. Extensive experiments on five popular MTS datasets verify the effectiveness of our proposed method.
\end{abstract}

\begin{links}
    \link{Code}{https://github.com/hdtkk/HDT}
\end{links}

\section{Introduction}

Multivariate time series forecasting task has been applied to many real-world applications, such as economics \cite{sezer2020financial, feng2022relation}, traffic \cite{wu2020adversarial, liu2016online}, energy \cite{zhichengsdformer} and weather \cite{qiu2017short, jin2023survey}. As a generative task, MTS forecasting presents challenges in two key aspects: the inherent high-dimensionality of the data distribution, and the long-term forecasting. To model the complex distributions of high-dimensional data, previous studies have established deep generative models in both autoregressive and non-autoregressive ways. To our knowledge, most of the work in the context of high-dimensional MTS has focused on short-term forecasting (predicted length: 24, 48) \cite{rasul2020multivariate,rasulvq, fan2024mg}.To improve long-term forecasting, various Transformer architectures \cite{nie2022time, liu2023itransformer} have been proposed, but most are focused on low-dimensional scenarios. Effectively modeling high-dimensional distributions with longer forecasting lengths remains a challenge. A key issue is integrating deep generative models with sequence modeling frameworks to handle both high-dimensional data and long-term forecasting tasks.

Existing works \cite{salinas2020deepar, rasul2021autoregressive, li2022generative, feng2023multi} have several attempts to utilize various forms of deep generative models, such as Normalizing flows \cite{dinh2016density}, Variational Auto-Encoder (VAEs) \cite{kingma2013auto}, Diffusion models \cite{litransformer, fan2024mg} to model high-dimensional MTS. They apply deep generative models to the high-dimensional distributions over time, learning the patterns of distribution changes along the temporal dimension for precise prediction. Due to complex patterns and long temporal dependencies of MTS, directly modeling high-dimensional MTS distributions in the time domain can lead to issues of distribution drift \cite{kim2021reversible} and overlook the correlations between variables, limited to short-term forecasting settings. 

Recently, several attention-variant Transformer frameworks  \cite{, liu2023itransformer, rao2022revisiting} and LLM-based structures \cite{zhou2023one, bian2024multi} have been applied to long-term forecasting of MTS, showing excellent performance on MTS datasets. Building on the success of these methods, we identified two key modules: the series decomposition block \cite{wu2021autoformer, liu2022non}, which uses moving averages to smooth periodic fluctuations and highlight long-term trends, and the discrete Transformer for MTS modeling. Inspired by these approaches, we first learn the discrete representations of the MTS and then incorporate the long-term trends of the forecasting target into our model. This allows us to enhance forecasting length capability with high accuracy.

As a discrete framework, Vector Quantized \cite{gray1984vector} techniques have shown strong competitiveness in high-dimensional image fields \cite{ijcai2021p0135, zheng2022movq, chang2023muse}, These approaches utilize the pre-quantizing images into discrete latent variables and modeling them autoregressively. For the time series domain, VQ-based methods such as TimeVQVAE \cite{lee2023vector}, TimeVAE \cite{desai2021timevae} and TimeGAN \cite{yoon2019time} all focus on time series generation task, the lateset VQ-TR \cite{rasulvq} introduce the VQ strategy within the transformer architecture as part of the encoder attention blocks, which attends over larger context windows with linear complexity in sequence length for efficient probabilistic forecasting. Inspired by their success of discrete strategy, we aim to explore the application of these techniques in the domain of high-dimensional MTS. Our model differs VQ-TR in two key aspects: i) HDT is two-stage, whereas it is end-to-end. ii) We focus on enhancing the long-term forecasting performance by introducing discrete representation of target itself, while they take efforts to reduce time and space complexity by discretizing the context inputs for efficient forecasting.

To extend the forecasting length within the high-dimensional MTS, we propose an effective generative framework, which is called Hierarchical Discrete Transformer \textbf{HDT}. It is a two-stage learning framework, consisting of a pre-quantizing module to obtain the discrete latent tokens of the forecasting targets, called tokenization, and a hierarchical modeling strategy for generating the discrete tokens. In the stage 1, we design two discrete token learning modules: one for obtaining latent tokens of our forecasting targets, and the other for obtaining latent tokens of downsampled targets using the downsampled input. This approach yields two key benefits: i) compressed latent discrete tokens effectively extend the prediction length for high-dimensional MTS, and ii) by incorporating the discrete latent space features of the targets, we reduce time complexity through shorter discrete token generation in stage 2.

In the stage 2, we devise a hierarchical discrete Transformer. At the low-level, we perform cross-attention between the contextual information and the discrete downsampled targets to generation task of downsampling target. At the high-level, we use the discrete downsampled results generated at the low-level as conditions to perform self-conditioned cross-attention with the discrete target, thereby achieving the generation of the discrete target. We summarize our main contributions as follows.
\begin{itemize}
\item We propose an effective hierarchical vector quantized method to introduce the long-term trend of targets for future target forecasting with higher accuracy and faster inference time.
\vspace{-0.05in}
\item We build a vector quantized MTS framework with $\ell_2$ normalization and self-conditioned cross attention for MTS forecasting, which can scale to high-dimensional and extend the prediction length with high accuracy.
\vspace{-0.05in}
\item Extensive experiments conducted on real-world datasets demonstrate the superiority of our \textbf{HDT}, achieving an average \textbf{16.7}\% improvement on $\text{CRPS}_{\text{sum}}$ and \textbf{15.4}\% on $\text{NRMSE}_{\text{sum}}$, compared to the state-of-the-art methods.
\end{itemize}

\begin{figure*}[t!]
  \centering
  \includegraphics[width=0.60\linewidth]{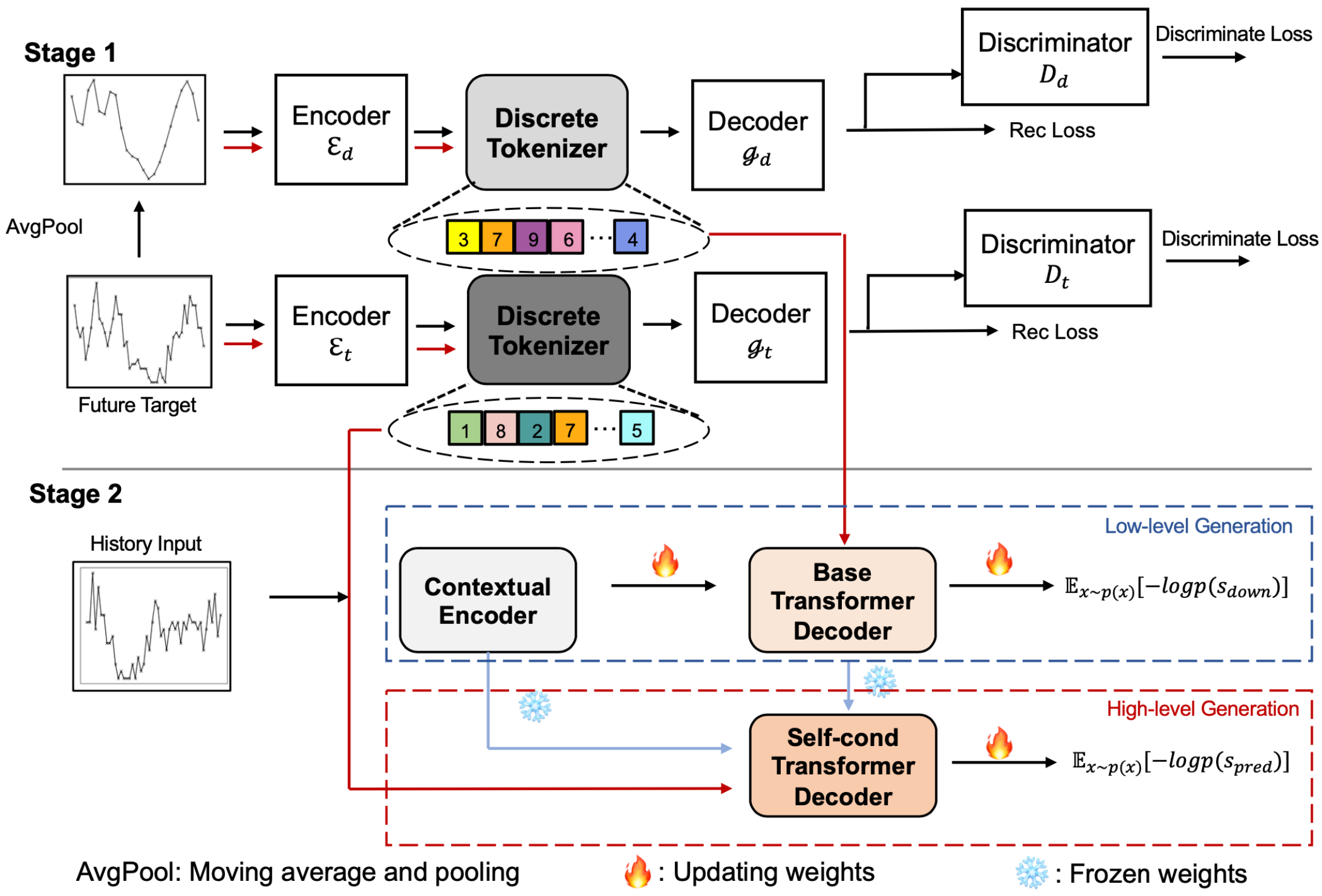}
  \caption{An illustration of our proposed HDT is provided. In stage 1, the model generates discrete downsampled targets and discrete targets, which are passed to Stage 2 for further processing. In stage 2, the contextual encoder and base Transformer decoder are trained with historical inputs and discrete downsampled tokens at the low level. Once trained, these low-level modules are fixed, and their outputs are fed into the high-level framework to generate the final discrete target sequence.}
  \label{overstructure}
\end{figure*} 

\section{Methods}
Our model comprises several key components. In this section, we present an overview of these components, which are divided into two stages. The training and inference details are shown in Algorithm~\ref{algorithm_1},~\ref{algorithm_2} and~\ref{algorithm_3}. Figure~\ref{overstructure} provides an overview of the model architecture. 
In the stage 1, we have two types of VQGAN \cite{esser2021taming} structures (Encoder, Quantization, Decoder): one is based on the discrete representation learning of the downsampled time series, and the other is based on the discrete representation learning structure corresponding to the prediction targets. Since the VQ strategy is operated on the channel dimension, the inter-variate correlations are captured in stage 1. In stage 2, a context encoder and a base Transformer decoder perform temporal cross-attention to generate discrete downsampled targets. The output from these low-level modules is then fed into a self-conditioned Transformer decoder to autoregressively predict discrete target tokens. This two-stage approach captures inter- and intra-correlations with discrete tokens, enhancing the accuracy of time series forecasting. 

\subsection{Stage 1: Modulating Quantized Vector}

\textbf{Series Downsample Module.} According to the Autoformer \cite{wu2021autoformer}, the moving average operation of non-stationary time series can smooth out periodic fluctuations and highlight long-term trends. As the objective of our work is to address the challenge of long-term forecasting in high-dimensional MTS, it is crucial for us to retain long-term patterns with the downsampled time series. For length-$\tau$ input series $\mathcal{X}_{pred} \in \mathbb{R}^{\tau \times D}$, the process is:
\begin{equation}
\mathcal{X}_{down}=\operatorname{AvgPool}(\operatorname{Padding}(\mathcal{X}_{pred})),
\label{11111}
\end{equation}
where $\mathcal{X}_{down} \in \mathbb{R}^{\tau \times D}$ denotes the long-term pattern representations. Here, we introduce the $\operatorname{AvgPool}(.)$ for moving average with the $\operatorname{Padding}(.)$ to keep the series length unchanged. $\mathcal{X}_{down}$ is the self-condition of targets, which consists of long-term patterns for the following future targets forecasting.

\noindent \textbf{Discrete Tokenization using VQGAN.} In the discrete representation learning of stage 1, the discrete learning modules of targets and downsampled targets show the same structure, which consists of an encoder and a decoder, with a quantization layer that maps a time series input into a sequence of tokens from a learned codebook. The details of these modules are provided in the Appendix C. Specifically, given any time series $\mathcal{X}_{pred} \in \mathbb{R}^{\tau \times D}$ can be represented by a spatial collection of codebook entries $z_{\textbf{q}_{t}} \in \mathbb{R}^{s \times n_{z}}$, where $n_{z}$ is the dimensionality of quantized vectors in the codebook and $s$ is the length of the discrete token sequence. In this way, each time series can be equivalently represented as a compact sequence with $s$ indices of the code vectors. The quantization operates on the channel dimension, capturing inter-variate correlations. Formally, the observed target $\mathcal{X}_{pred}$ and downsampled target $\mathcal{X}_{down}$ are reconstructed by:
\begin{align}
\hat{\mathcal{X}_{pred}}&=\mathcal{G}_{\theta_{t}}\left(z_{\textbf{q}_{t}}\right)=\mathcal{G}_{\theta_{t}}(\mathbf{q}_{t}(\hat{z}^{t}))=\mathcal{G}_{\theta_{t}}\left(\mathbf{q}_{t}\left(\mathcal{E}_{\psi_{t}}(\mathcal{X}_{pred})\right)\right), \\
\hat{\mathcal{X}_{down}}&=\mathcal{G}_{\theta_{d}}\left(z_{\textbf{q}_{d}}\right)=\mathcal{G}_{\theta_{d}}(\mathbf{q}_{d}(\hat{z}^{d}))=\mathcal{G}_{\theta_{d}}\left(\mathbf{q}_{d}\left(\mathcal{E}_{\psi_{d}}(\mathcal{X}_{down})\right)\right). 
\label{22222}
\end{align}
In particular, the $\mathcal{E}_{\psi_{[t, d]}}, \textbf{q}_{[t, d]},  \mathcal{G}_{\theta_{[t,d]}}$ are the encoders, quantization layers and decoders corresponding to $\mathcal{X}_{pred}$ and $\mathcal{X}_{down}$, respectively. To avoid confusion and redundant expressions, we have removed the subscript symbols corresponding to the discrete learning and training process in the stage 1 formulas. The quantization operator $\mathbf{q}$ is  conducted to transfer the continuous feature into the discrete space by looking up the closest codebook entry $z_{k}$ for each timestamp feature $\hat{z_{i}}$ within $\hat{z}$, and note that $\hat{z}$ represents the execution process corresponding to both $\hat{z}^{t}$ and $\hat{z}^{d}$.:
\begin{equation}
z_q=\mathbf{q}(\hat{z})=\underset{z_k \in \mathcal{Z}}{\arg \min }\left\|\hat{z}_{i}-z_k\right\|,
\label{33333}
\end{equation}
where $\mathcal{Z} \in \mathbb{R}^{K \times n_{z}}$ is the codebook that consists of $K$ entries with $n_{z}$ dimensions and $\hat{z_{i}}$ is the continuous feature of the timestamp. Note that $z_{\textbf{q}_{t}}$ and $z_{\textbf{q}_{d}}$ each correspond to their respective codebooks $\mathcal{Z}^{t}$ and $\mathcal{Z}^{d}$. The subscript for $\mathcal{Z}$ is omitted to maintain the brevity of the paper. The above models and the codebook can be learned by optimizing the following objectives:
\begin{align}
\mathcal{L}_{VQ}\left(\mathcal{E}_\psi, \mathcal{G}_\theta, \mathcal{Z}\right) = & \|\mathcal{X}-\hat{\mathcal{X}}\|_2^2 + \left\|\operatorname{sg}\left[\mathcal{E}_\psi(\mathcal{X})\right]-z_q\right\|_2^2 \nonumber \\
& + \beta\left\|\operatorname{sg}\left[z_q\right]-\mathcal{E}_\psi(\mathcal{X})\right\|_2^2.
\label{4444}
\end{align}
In detail, $\operatorname{sg}$ denotes the stop-gradient operator, $\beta$ is a hyperparameter for the last term $commitment$ $loss$. The first term is $reconstruction$ $loss$ and the second is $codebook$ $ loss$ to optimize the entries in the codebook.

To learn a perceptually rich codebook in VQGAN, it introduces an adversarial training procedure with a patch-based discriminator $D$=\{$D_{t}, D_{d}$\} \cite{isola2017image} that aims to differentiate between real and reconstructed images. In our setting, we introduce a shallow Conv1d network to enhance the reconstruction results:
\begin{equation}
\mathcal{L}_{\mathrm{GAN}}(\{\mathcal{E}_\psi, \mathcal{G}_\theta,\mathcal{Z}\}, D)=[\log D(\mathcal{X})+\log (1-D(\hat{\mathcal{X}}))].
\label{5555}
\end{equation}
The final objective for finding the optimal Model $\mathcal{Q}^*={\mathcal{E}_\psi, \mathcal{G}_\theta, \mathcal{Z}}$ is:
\begin{align*}
\mathcal{Q}^* &={\arg\min_{\mathcal{E}_\psi,\mathcal{G}_\theta, \mathcal{Z}}} \max _D \mathbb{E}_{\mathcal{X} \sim p(\mathcal{X})}\left[\mathcal{L}_{\mathrm{VQ}}(\mathcal{E}_\psi, \mathcal{G}_\theta,\mathcal{Z})\right. \nonumber \\
& + \left.\lambda \mathcal{L}_{\mathrm{GAN}}(\{\mathcal{E}_\psi, \mathcal{G}_\theta, \mathcal{Z}\}, D)\right],
\label{6666}
\end{align*}
where the $\lambda$ is an adaptive weight parameter, which is computed by the gradient of $\mathcal{G}_{\theta}$ and $D$.

\noindent $\ell_2$ \textbf{Regularization.} However, in our experiments, we observed that applying $\mathbf{l}_{2}$ normalization ($\frac{\mathbf{x}}{\|\mathbf{x}\|_2}$) to the entries in the codebook can enhance the reconstruction performance.
\begin{equation}
\mathcal{L}_{norm} = \left\|\ell_2\left(\mathcal{E}_\psi(\mathcal{X})\right)-\ell_2\left(z_{k}\right)\right\|_2^2.
\label{7777}
\end{equation}
Finally, the training loss function is described as:
\begin{equation}
\mathcal{L} = \mathcal{L}_{VQ}\left(\mathcal{E}_\psi, \mathcal{G}_\theta, \mathcal{Z}\right) + \mathcal{L}_{\mathrm{GAN}}(\{\mathcal{E}_\psi, \mathcal{G}_\theta,\mathcal{Z}\}, D) + \mathcal{L}_{norm}.
\label{8888}
\end{equation}
Overall, in the stage 1, $\mathcal{X}_{pred}$ and $\mathcal{X}_{down}$  each obtain their respective codebooks $\mathcal{Z}^{t}$ and $\mathcal{Z}^{d}$.
\subsection{Stage 2: Modelling Prior Distribution with HDT} In this section, we introduce the details of the hierarchical discrete transformer. 
In stage 2, we establish a framework to estimate the underlying prior distribution over the discrete space for generating discrete time series tokens. This allows the post-quantization layers and the decoder from stage 1 to reconstruct the continuous targets. First, we present the overall generation process for the discrete tokens, as illustrated in Figure 1. Then, we detail the specific implementation procedures for both the low-level and high-level generation separately. \\
\textbf{Low-level Token Generation.} This process can be considered a preliminary process of target token generation of high-level. Specifically, we now have the context data $\mathcal{X}_{p} \in \mathbb{R}^{h \times D}$ and the discrete representation of the downsampled target $s_{down}=\{z^{s_{1}}_{\textbf{q}_{d}}, z^{s_{2}}_{\textbf{q}_{d}},..., z^{s_{d}}_{\textbf{q}_{d}} \} \in \mathbb{R}^{s_{d} \times n_{z}}$, where h is the look-back window length and $D$ is the number of variates, $s_d$ is the length of discrete downsampled target sequence and $n_{z}$ is the feature dimension of the discrete representation. We formulate the training process by:
\begin{align}
& \mathcal{H}_{p} =\mathcal{E}_{T}(\mathcal{X}_{p}), \\
& p(s_{down}|c) =\prod_i p\left(z^{s_{i}}_{\textbf{q}_{d}} \mid z^{s_{<i}}_{\textbf{q}_{d}}, c=\mathcal{H}_{p}\right),\\
& \mathcal{L}_{\operatorname{base}} =\mathbb{E}_{x \sim p(x)}[-\log p(s_{down})],  
\label{1212}    
\end{align}
where $\mathcal{E}_{T}$ is the contextual encoder that is the Transformer encoder in our experiment. $\mathcal{H}_{p} \in  \mathbb{R}^{h \times n_{z}}$ is the output of the context encoder and $ \mathcal{L}_{\operatorname{base}}$ is the loss function of base Transformer decoder at the low-level framework.  $p\left(z^{s_{i}}_{\textbf{q}_{d}} \mid z^{s_{<i}}_{\textbf{q}_{d}}, c=\mathcal{H}_{p}\right)$ is to compute the likelihood of the full representation $p(s_{down}|c)= \prod_i p\left(z^{s_{i}}_{\textbf{q}_{d}} \mid z^{s_{<i}}_{\textbf{q}_{d}}, c=\mathcal{H}_{p}\right).$ We then obtain the trained context embedding $\mathcal{H}_{p}$ and the downsampled tokens $s_{down}$. Moreover, the discrete downsampled results directly impact the generation of high-level discrete targets, we explored three different methods for obtaining $\mathcal{H}_{p}$. These methods are explained in detail in the subsequent experimental section.\\
\textbf{High-level Token Generation.} After training the context encoder and base Transformer decoder in the low-level framework, we not only capture the content features of the context but also ensure that the discrete downsampled sequences retain long-term patterns. This provides additional conditions related to the target's own features in the high-level framework, thereby enhancing the accuracy of long-term forecasting. We have the discrete target $s_{pred}=\{z^{s_{1}}_{\textbf{q}_{t}}, z^{s_{2}}_{\textbf{q}_{t}},..., z^{s_{p}}_{\textbf{q}_{t}} \} \in \mathbb{R}^{s_{p} \times n_{z}}$, $s_{down}$ and $\mathcal{H}_{p}$, where the $s_{p}$ is the length of discrete target sequence. The process of autoregressively generating $s_{pred}$ can be described as follows:
\begin{align}
& p(s_{pred}|c) =\prod_i p\left(z^{s_{i}}_{\textbf{q}_{t}} \mid  z^{s_{<i}}_{\textbf{q}_{t}}, c=\{s_{down}, \mathcal{H}_{p}\} \right), \\
& \mathcal{L}_{\operatorname{self-cond}} =\mathbb{E}_{x \sim p(x)}[-\log p(s_{pred})],  
\label{selfcond_cond} 
\end{align}
where the $s_{down}$ and $\mathcal{H}^{p}$ are fixed, the cross-attention of self-conditioned Transformer decoder is operating between the $s_{down}$ and $s_{pred}$, the temporal cross-attention is introduced to the $\mathcal{H}^{p}$ and $s_{pred}$, as shown in Figure~\ref{overstructure}. After completing the high-level training, we can input the discrete form of the target into the stage 1 decoder $\mathcal{G}_{\theta_{t}}$ to reconstruct the predicted target. Notably, unlike the popular diffusion models, the VQ discretization strategy effectively avoids the efficiency issues associated with iterative diffusion structures and autoregressive prediction methods.

\begin{algorithm}[htbp]
    \caption{Training of Stage 1}\label{algorithm_1}
     \textbf{Input:} Set of time series targets  $\mathcal{X}_{pred}$\\ 
    \textbf{Output:} Encoder $\mathcal{E}_{\psi_{t}}$ and $\mathcal{E}_{\psi_{d}}$, Decoder $\mathcal{G}_{\theta_{t}}$ and $\mathcal{G}_{\theta_{d}}$, Discriminator $D_{t}$ and $D_{d}$, quantization codebook $\mathbf{q}_{t}$ and $\mathbf{q}_{d}$. \\
    \vspace{-0.2in}
    \begin{algorithmic}[1]
    \FOR{$k\leftarrow 1$ to $K$}
        \STATE Get the $\text{X}_{\text{pred}} \sim \mathcal{X}_{pred}$; \\
        \STATE Obtain the $\text{X}_{\text{down}}$ by Eqn.~\ref{11111};\\
        \STATE Feed $\text{X}_{\text{pred}}$ and $\text{X}_{\text{down}}$ to encoder \{$\mathcal{E}_{\psi_{t}}$, $\mathcal{E}_{\psi_{d}}$\}, and quantization \{$\mathbf{q}_{t}$, $\mathbf{q}_{d}$\}, by Eqn.(2, 3, 4), respectively; \\
        \STATE Compute the $\ell_2$ Regularization and loss by Eqn.(~\ref{4444},~\ref{7777}); \\
        \IF {k $\geq$ $\hat{k}$ is 0.75$K$}
            \STATE Introduce the Discriminator $D_{t}$, $D_{d}$ respectively and compute the loss by Eqn.~\ref{8888};
        \ENDIF
    \ENDFOR
    \STATE Return trained $\mathcal{E}_{\psi_{t}}$, $\mathcal{E}_{\psi_{d}}$, $\mathcal{G}_{\theta_{t}}$, $\mathcal{G}_{\theta_{d}}$, $\mathbf{q}_{t}$, $\mathbf{q}_{d}$, $D_{t}$ and $D_{d}$.\\
    \end{algorithmic}
\end{algorithm}
\begin{algorithm}[htbp]
    \caption{Training of Stage II}\label{algorithm_2}
    \textbf{Input:} Set of history time series $\mathcal{X}_{p}$, targets $X_{pred}$ and trainable BOS token [BOS]. The optimized encoders $\mathcal{E}_{\psi_{d}}$ and $\mathcal{G}_{\theta_{t}}$, trained quantization codebooks $\mathbf{q}_{t}$ and $\mathbf{q}_{d}$.  \\ 
    \textbf{Output:} The base Transformer decoder $\mathcal{B}$, contextual encoder $\mathcal{E}_{T}$, and self-cond Transformer decoder$\mathcal{S}$. \\
    \vspace{-0.2in}
    \begin{algorithmic}[1]
    \FOR{$k\leftarrow 1$ to $K$}
        \STATE Obtain the $X_{down}$ from $X_{pred}$by Eqn.~\ref{11111};
        \STATE Get the token sequences $s_{down}$ and $s_{pred}$ from trained $\mathbf{q}_{t}$ and $\mathbf{q}_{d}$ of stage 1 by Eqn.~\ref{33333} with $X_{down}$ and $X_{pred}$, respectively; \\
        \STATE Minimize the negative log-likelihood with training $\mathcal{E}_{T}$ and $\mathcal{B}$ by Eqn.(9, 10, 11) with concatenating the [BOS] token at the beginning of token sequence $s_{down}$. \\
    \ENDFOR   
    \FOR{$k\leftarrow 1$ to $K$}
            \STATE Introduce the output $s_{down}$ from the combination of trained $\mathcal{E}_{T}$ and $\mathcal{B}$;\\
            \STATE Minimize the negative log-likelihood with frozen $\mathcal{E}_{T}$, $\mathcal{B}$ and trainable $\mathcal{S}$ by Eqn.~\ref{selfcond_cond} with concatenating the [BOS] token at the beginning of token sequence $s_{pred}$. \\
    \ENDFOR
    \STATE Return trained contextual encoder $\mathcal{E}_{T}$, base Transformer decoder $\mathcal{B}$, self-cond Transformer decoder $\mathcal{S}$ and [BOS] token.\\
    \end{algorithmic}
\end{algorithm}

\begin{algorithm}[ht]
    \caption{Inference}\label{algorithm_3}
    \textbf{Input:} Set of history time series $\mathcal{X}_{p}$, trained BOS token [BOS], trained contextual encoder $\mathcal{E}_{T}$, base Transformer decoder $\mathcal{B}$ and self-cond Transformer $\mathcal{S}$ and Decoder $\mathcal{G}_{\theta_{t}}$.  \\ 
    \textbf{Output:} Reconstructed future targets  $\mathcal{X}_{pred}$. \\
    \vspace{-0.2in}
    \begin{algorithmic}[1]
    \FOR{$i\leftarrow 1$ to $I$ in test samples}
        \STATE Sample the downsampled tokens $s_{down}$ with trained [BOS] token and $\mathcal{X}_{p}$ from the combination of $\mathcal{E}_{T}$ and $\mathcal{B}$ by Eqn.(9, 10, 11); \\
        \STATE Sample the target tokens $s$ with [BOS] token from trained $\mathcal{S}$, $\mathcal{E}_{T}$ and $\mathcal{B}$ by Eqn.(12, 13); \\
        \STATE Return the target $X_{pred}$ by $\mathcal{G}_{\theta_{t}}$ of Eqn.2.
    \ENDFOR
    \STATE Return the prediction target $\mathcal{X}_{pred}$. \\
    \end{algorithmic}
\end{algorithm}

\section{Experiments}
We conducted experiments to evaluate the performance and efficiency of HDT, covering short-term and long-term forecasting as well as robustness to missing values. The evaluation includes 5 real-world benchmarks and 12 baselines. Detailed model and experiment configurations are summarized in Appendix C. \\
\textbf{Datasets.} We extensively evaluate the proposed HDT on five real-world benchmarks, covering the mainstream high-dimensional MTS probabilistic forecasting applications, Solar \cite{lai2018modeling}, Electricity \cite{lai2018modeling}, Traffic \cite{salinas2019high}, Taxi \cite{salinas2019high} and Wikipedia \cite{gasthaus2019probabilistic}. These data are recorded at intervals of 30 minutes, 1 hour, and 1 day frequencies, more details refer to Appendix B.\\
\textbf{Baselines.} We include several competitive multivariate time series baselines to verify the effectiveness of HDT. Previous work DeepAR \cite{salinas2020deepar}, GP-Copula \cite{salinas2019high} and Transformer-MANF \cite{rasul2020multivariate}. Then, we compare HDT against the diffusion-based methods, TimeGrad \cite{rasul2021autoregressive}, MG\_TSD \cite{fan2024mg}, $\operatorname{D}^{3}$VAE \cite{li2022generative}, CSDI \cite{tashiro2021csdi}, SSSD \cite{alcaraz2022diffusion}, TSDiff \cite{kollovieh2023predict} with additional Transformer layers followed by S4 layer and TimeDiff \cite{shen2023non}. Among the MTS forecasting with VQ-Transformer, we introduce and VQ-TR \cite{rasul2023vq} for comparisons. The details of baselines are shown in Appendix F.\\
\textbf{Evaluation Metrics.} For probabilistic estimates, we report the continuously ranked probability score across summed time series $(\text{CRPS}_{sum})$ \cite{matheson1976scoring}, a widely used metric for probabilistic time series forecasting, as well as a deterministic estimation metric $\text{NRMSE}_{sum}$ (Normalized Root Mean Squared Error). For detailed descriptions, refer to Appendix B.\\
\textbf{Implementation Details.} Our method relies on the ADAM optimizer with initial learning rates of 0.0005 and 0.001, and a batch size of 64 across all datasets. The history length is fixed at 96, with prediction lengths of \(\{48, 96, 144\}\). We sample 100 times to report metrics on the test set. All experiments are conducted on a single Nvidia A-100 GPU, and results are based on 3 runs.

\begin{table*}[t]
  \vskip -0.0in
  \vspace{3pt}
  \renewcommand{\arraystretch}{0.85} 
  \centering
  \resizebox{1.0\linewidth}{!}{
  \begin{threeparttable}
  \begin{small}
  \renewcommand{\multirowsetup}{\centering}
  \setlength{\tabcolsep}{1pt}
  \begin{tabular}{c|c|cc|cc|cc|cc|cc|cc|cc|cc|cc|cc|cc|cc}
    \toprule
    \multicolumn{2}{c}{\multirow{2}{*}{Models}} & 
    \multicolumn{2}{c}{\rotatebox{0}{\scalebox{0.8}{\textbf{HDT}}}} &
    \multicolumn{2}{c}{\rotatebox{0}{\scalebox{0.8}{\update{VQ-TR}}}} &
    \multicolumn{2}{c}{\rotatebox{0}{\scalebox{0.8}{MG\_TSD}}} &
    \multicolumn{2}{c}{\rotatebox{0}{\scalebox{0.8}{TSDiff}}}  &
    \multicolumn{2}{c}{\rotatebox{0}{\scalebox{0.8}{TimeDiff}}}  &
    \multicolumn{2}{c}{\rotatebox{0}{\scalebox{0.8}{SSSD}}} &
    \multicolumn{2}{c}{\rotatebox{0}{\scalebox{0.8}{{$\operatorname{D}^{3}$VAE}}}} &
    \multicolumn{2}{c}{\rotatebox{0}{\scalebox{0.8}{CSDI}}}&
    \multicolumn{2}{c}{\rotatebox{0}{\scalebox{0.8}{TimeGrad}}} &
    \multicolumn{2}{c}{\rotatebox{0}{\scalebox{0.8}{Trans-MAF}}} &
    \multicolumn{2}{c}{\rotatebox{0}{\scalebox{0.8}{DeepAR}}} 
    &\multicolumn{2}{c}{\rotatebox{0}{\scalebox{0.8}{GP-Copula}}} \\
    \multicolumn{2}{c}{} &
    \multicolumn{2}{c}{\scalebox{0.8}{\textbf{(Ours)}}} & 
    \multicolumn{2}{c}{\scalebox{0.8}{(2024)}} & 
    \multicolumn{2}{c}{\scalebox{0.8}{\citeyearpar{fan2024mg}}} & 
    \multicolumn{2}{c}{\scalebox{0.8}{\citeyearpar{kollovieh2024predict}}}  & 
    \multicolumn{2}{c}{\scalebox{0.8}{\citeyearpar{shen2023non}}}  & 
    \multicolumn{2}{c}{\scalebox{0.8}{(2023)}} & 
    \multicolumn{2}{c}{\scalebox{0.8}{\citeyearpar{li2022generative}}} & 
    \multicolumn{2}{c}{\scalebox{0.8}{\citeyearpar{tashiro2021csdi}}}& 
    \multicolumn{2}{c}{\scalebox{0.8}{\citeyearpar{rasul2021autoregressive}}} &
    \multicolumn{2}{c}{\scalebox{0.8}{\citeyearpar{rasul2020multivariate}}} &
    \multicolumn{2}{c}{\scalebox{0.8}{\citeyearpar{salinas2020deepar}}} 
    &\multicolumn{2}{c}{\scalebox{0.8}{\citeyearpar{salinas2019high}}} \\
    \cmidrule(lr){3-4} \cmidrule(lr){5-6}\cmidrule(lr){7-8} \cmidrule(lr){9-10}\cmidrule(lr){11-12}\cmidrule(lr){13-14} \cmidrule(lr){15-16} \cmidrule(lr){17-18} \cmidrule(lr){19-20} \cmidrule(lr){21-22} \cmidrule(lr){23-24} \cmidrule(lr){25-26}
    \multicolumn{2}{c}{Metric}  & \scalebox{0.78}{$\hat{\operatorname{CRPS}}$} & \scalebox{0.78}{$\hat{\text{NRMSE}}$}  & \scalebox{0.78}{$\hat{\operatorname{CRPS}}$} & \scalebox{0.78}{$\hat{\text{NRMSE}}$}  & \scalebox{0.78}{$\hat{\operatorname{CRPS}}$} & \scalebox{0.78}{$\hat{\text{NRMSE}}$}  & \scalebox{0.78}{$\hat{\operatorname{CRPS}}$} & \scalebox{0.78}{$\hat{\text{NRMSE}}$}  & \scalebox{0.78}{$\hat{\operatorname{CRPS}}$} & \scalebox{0.78}{$\hat{\text{NRMSE}}$}  & \scalebox{0.78}{$\hat{\operatorname{CRPS}}$} & \scalebox{0.78}{$\hat{\text{NRMSE}}$} & \scalebox{0.78}{$\hat{\operatorname{CRPS}}$} & \scalebox{0.78}{$\hat{\text{NRMSE}}$} & \scalebox{0.78}{$\hat{\operatorname{CRPS}}$} & \scalebox{0.78}{$\hat{\text{NRMSE}}$} & \scalebox{0.78}{$\hat{\operatorname{CRPS}}$} & \scalebox{0.78}{$\hat{\text{NRMSE}}$} & \scalebox{0.78}{$\hat{\operatorname{CRPS}}$} & \scalebox{0.78}{$\hat{\text{NRMSE}}$} & \scalebox{0.78}{$\hat{\operatorname{CRPS}}$} & \scalebox{0.78}{$\hat{\text{NRMSE}}$} & \scalebox{0.78}{$\hat{\operatorname{CRPS}}$} & \scalebox{0.78}{$\hat{\text{NRMSE}}$}\\
    \toprule
    \multirow{4}{*}{\update{\rotatebox{90}{\scalebox{0.95}{Solar}}}}
    &  \scalebox{0.78}{48} & \scalebox{0.78}{0.329} & \scalebox{0.78}{0.653} & \scalebox{0.78}{0.334} & \scalebox{0.78}{0.657} & \scalebox{0.78}{\underline{0.328}} & \scalebox{0.78}{\textbf{0.645}} &     \scalebox{0.78}{\textbf{0.324}} & \scalebox{0.78}{\underline{0.651}} & \scalebox{0.78}{0.376} & \scalebox{0.78}{0.814} & \scalebox{0.78}{0.340} & \scalebox{0.78}{0.654} &{\scalebox{0.78}{0.382}} &{\scalebox{0.78}{0.692}} &{\scalebox{0.78}{0.336}} &{\scalebox{0.78}{0.651}} & \scalebox{0.78}{0.357} & \scalebox{0.78}{0.667} &\scalebox{0.78}{0.341} &\scalebox{0.78}{0.672} &\scalebox{0.78}{0.362} &\scalebox{0.78}{0.691} &\scalebox{0.78}{0.426} &\scalebox{0.78}{0.891}\\ 
    & \scalebox{0.78}{96} & \scalebox{0.78}{\textbf{0.330}} & \scalebox{0.78}{\textbf{0.694}} & \scalebox{0.78}{0.357} & \scalebox{0.78}{0.734} & \scalebox{0.78}{0.339} & \scalebox{0.78}{0.707} & \scalebox{0.78}{\underline{0.336}} & \scalebox{0.78}{0.715} &\scalebox{0.78}{0.415} & \scalebox{0.78}{0.935} &\scalebox{0.78}{0.365} &\scalebox{0.78}{\underline{0.704}} &{\scalebox{0.78}{0.413}} &{\scalebox{0.78}{0.757}} & \scalebox{0.78}{0.359} & \scalebox{0.78}{0.712}  &\scalebox{0.78}{0.384} &\scalebox{0.78}{0.731} &\scalebox{0.78}{0.376} &\scalebox{0.78}{0.743}  &\scalebox{0.78}{0.402} &\scalebox{0.78}{0.775} &\scalebox{0.78}{0.475} &\scalebox{0.78}{0.921}\\ 
    & \scalebox{0.78}{144} & \scalebox{0.78}{\textbf{0.357}} & \scalebox{0.78}{\textbf{0.776}} & \scalebox{0.78}{0.377} & \scalebox{0.78}{0.885} & {\scalebox{0.78}{\underline{0.373}}} & \scalebox{0.78}{\underline{0.825}} & \scalebox{0.78}{0.379}  &\scalebox{0.78}{0.847} & \scalebox{0.78}{0.438} & \scalebox{0.78}{1.312} &\scalebox{0.78}{0.392} &\scalebox{0.78}{0.830}  &{\scalebox{0.78}{0.448}} &{\scalebox{0.78}{0.914}} & \scalebox{0.78}{0.387} & \scalebox{0.78}{0.865}  &\scalebox{0.78}{0.429} &\scalebox{0.78}{0.916} &\scalebox{0.78}{0.394} &\scalebox{0.78}{0.824} &\scalebox{0.78}{0.448} &\scalebox{0.78}{0.936} &\scalebox{0.78}{0.559} &\scalebox{0.78}{1.207} \\ 
    \cmidrule(lr){2-26}
    & \scalebox{0.78}{Avg} & \scalebox{0.78}{\textbf{0.338}} & \scalebox{0.78}{\textbf{0.707}} & \scalebox{0.78}{0.356} & \scalebox{0.78}{0.758} & \scalebox{0.78}{0.347} & \scalebox{0.78}{\underline{0.726}} & \scalebox{0.78}{\underline{0.346}} & \scalebox{0.78}{0.738} & \scalebox{0.78}{0.410} & \scalebox{0.78}{1.020} &\scalebox{0.78}{0.366} &\scalebox{0.78}{0.729}  &{\scalebox{0.78}{0.414}} &{\scalebox{0.78}{0.788}} & \scalebox{0.78}{0.361} & \scalebox{0.78}{0.743}  &\scalebox{0.78}{0.390} &\scalebox{0.78}{0.771} &\scalebox{0.78}{0.370} &\scalebox{0.78}{0.746} &\scalebox{0.78}{0.404} &\scalebox{0.78}{0.801}
    &\scalebox{0.78}{0.487} &\scalebox{0.78}{1.006} \\ 
    \midrule
    
    \multirow{4}{*}{\update{\rotatebox{90}{\scalebox{0.95}{Electricity}}}}
    &  \scalebox{0.78}{48} & \scalebox{0.78}{0.025} & \scalebox{0.78}{\underline{0.030}} & \scalebox{0.78}{0.034} & \scalebox{0.78}{0.033} & \scalebox{0.78}{\textbf{0.023}} & \scalebox{0.78}{\underline{0.030}} & \scalebox{0.78}{\underline{0.024}} & \scalebox{0.78}{\textbf{0.029}} & \scalebox{0.78}{0.036} & \scalebox{0.78}{0.092} &{\scalebox{0.78}{0.037}} &\scalebox{0.78}{0.032} &\scalebox{0.78}{0.046} &\scalebox{0.78}{0.096} & \scalebox{0.78}{0.032} & \scalebox{0.78}{0.034} &\scalebox{0.78}{0.043} &\scalebox{0.78}{0.031} &{\scalebox{0.78}{0.039}} &\scalebox{0.78}{0.034} &\scalebox{0.78}{0.043} &\scalebox{0.78}{0.035} &\scalebox{0.78}{0.047} &\scalebox{0.78}{0.055} \\ 
    & \scalebox{0.78}{96} & \scalebox{0.78}{\textbf{0.028}} & \scalebox{0.78}{\textbf{0.032}} & \scalebox{0.78}{0.045} & \scalebox{0.78}{0.040} & \scalebox{0.78}{\underline{0.034}} & \scalebox{0.78}{\underline{0.035}} & \scalebox{0.78}{0.039} & \scalebox{0.78}{0.036} & \scalebox{0.78}{0.049} & \scalebox{0.78}{0.109} &{\scalebox{0.78}{0.045}} &{\scalebox{0.78}{0.041}} &\scalebox{0.78}{0.062} &\scalebox{0.78}{0.114} & \scalebox{0.78}{0.049} & \scalebox{0.78}{0.039} &\scalebox{0.78}{0.067} &\scalebox{0.78}{0.035} &\scalebox{0.78}{0.060} &\scalebox{0.78}{0.038} &\scalebox{0.78}{0.058} &\scalebox{0.78}{0.044} &\scalebox{0.78}{0.069} &\scalebox{0.78}{0.058} \\ 
    & \scalebox{0.78}{144} & \scalebox{0.78}{\textbf{0.036}} & \scalebox{0.78}{\textbf{0.057}} & \scalebox{0.78}{0.049} & \scalebox{0.78}{0.070} & \scalebox{0.78}{\underline{0.042}} & \scalebox{0.78}{\underline{0.064}}  & \scalebox{0.78}{0.047} & \scalebox{0.78}{0.072}  & \scalebox{0.78}{0.063} & \scalebox{0.78}{0.147} &{\scalebox{0.78}{0.056}} &{\scalebox{0.78}{0.084}} &\scalebox{0.78}{0.086} &\scalebox{0.78}{0.142} & \scalebox{0.78}{0.067} & \scalebox{0.78}{0.088 } &\scalebox{0.78}{0.082} &\scalebox{0.78}{0.085} &\scalebox{0.78}{0.101} &\scalebox{0.78}{0.093} &\scalebox{0.78}{0.104} &\scalebox{0.78}{0.097} &\scalebox{0.78} {0.125} &\scalebox{0.78}{0.109} \\ 
    \cmidrule(lr){2-26}
    & \scalebox{0.78}{Avg} & \scalebox{0.78}{\textbf{0.028}} & \scalebox{0.78}{\textbf{0.038}} & \scalebox{0.78}{0.043} & \scalebox{0.78}{0.048} & \scalebox{0.78}{\underline{0.033}} & \scalebox{0.78}{\underline{0.043}} & \scalebox{0.78}{0.036} & \scalebox{0.78}{0.046} & \scalebox{0.78}{0.049} & \scalebox{0.78}{0.116} &{\scalebox{0.78}{0.046}} &{\scalebox{0.78}{0.052}} &\scalebox{0.78}{0.065} &\scalebox{0.78}{0.117} & \scalebox{0.78}{0.049} & \scalebox{0.78}{0.054} &\scalebox{0.78}{0.064} &\scalebox{0.78}{0.050} &\scalebox{0.78}{0.066} &\scalebox{0.78}{0.055} &\scalebox{0.78}{0.068} &\scalebox{0.78}{0.059}  &\scalebox{0.78}{0.080} &\scalebox{0.78}{0.074} \\
    \midrule
    
    \multirow{4}{*}{\rotatebox{90}{\update{\scalebox{0.95}{Traffic}}}}
    &  \scalebox{0.78}{48} & \scalebox{0.78}{\textbf{0.034}} & \scalebox{0.78}{\textbf{0.060}} & \scalebox{0.78}{0.039} & \scalebox{0.78}{0.074} & \scalebox{0.78}{\underline{0.036}} & \scalebox{0.78}{\underline{0.067}} & \scalebox{0.78}{0.057} & \scalebox{0.78}{0.070} & \scalebox{0.78}{0.064}& \scalebox{0.78}{0.175}  &\scalebox{0.78}{0.053} &{\scalebox{0.78}{0.074}} & \scalebox{0.78}{0.082} &\scalebox{0.78}{0.312}& \scalebox{0.78}{-} & \scalebox{0.78}{-} &\scalebox{0.78}{0.067} &\scalebox{0.78}{0.072} &\scalebox{0.78}{0.070} &\scalebox{0.78}{0.074} &\scalebox{0.78}{0.069} &\scalebox{0.78}{0.081} 
    &\scalebox{0.78}{0.082} &\scalebox{0.78}{0.136} \\
    
    & \scalebox{0.78}{96} & \scalebox{0.78}{\textbf{0.037}} & \scalebox{0.78}{\textbf{0.063}} & {\scalebox{0.78}{0.052}} & \scalebox{0.78}{0.082} & \scalebox{0.78}{\underline{0.042}} & \scalebox{0.78}{\underline{0.072}} & \scalebox{0.78}{0.068} & \scalebox{0.78}{0.076}  & \scalebox{0.78}{0.081} & \scalebox{0.78}{0.246} &\scalebox{0.78}{0.069} &\scalebox{0.78}{0.080} &{\scalebox{0.78}{0.091 }} &{\scalebox{0.78}{0.465}} & \scalebox{0.78}{-} & \scalebox{0.78}{-} &\scalebox{0.78}{0.095} &\scalebox{0.78}{0.087} &\scalebox{0.78}{0.086} &\scalebox{0.78}{0.081} &\scalebox{0.78}{0.099} &\scalebox{0.78}{0.128}  &\scalebox{0.78}{0.093} &\scalebox{0.78}{0.148} \\
    & \scalebox{0.78}{144} & \scalebox{0.78}{\textbf{0.047}} & \scalebox{0.78}{\textbf{0.076}} & \scalebox{0.78}{0.068} & \scalebox{0.78}{\underline{0.093}} & \scalebox{0.78}{\underline{0.056}} & \scalebox{0.78}{0.096} & \scalebox{0.78}{0.095} & \scalebox{0.78}{0.114} & \scalebox{0.78}{0.109 } & \scalebox{0.78}{0.304} &\scalebox{0.78}{0.084} &\scalebox{0.78}{0.106} &{\scalebox{0.78}{0.129}} & {\scalebox{0.78}{0.472}} & \scalebox{0.78}{-} & \scalebox{0.78}{-} &\scalebox{0.78}{0.124} &{\scalebox{0.78}{0.105}} &\scalebox{0.78}{0.107} &\scalebox{0.78}{0.096} &\scalebox{0.78}{0.113} &\scalebox{0.78}{0.142} 
    &\scalebox{0.78}{0.125} &\scalebox{0.78}{0.185} \\
    \cmidrule(lr){2-26}
    & \scalebox{0.78}{Avg} & \scalebox{0.78}{\textbf{0.039}} & \scalebox{0.78}{\textbf{0.066}} & \scalebox{0.78}{0.053} & \scalebox{0.78}{0.083} & \scalebox{0.78}{\underline{0.045}} & \scalebox{0.78}{\underline{0.078}} & \scalebox{0.78}{0.073} & \scalebox{0.78}{0.087} & \scalebox{0.78}{0.085} & \scalebox{0.78}{0.241} &\scalebox{0.78}{0.069} &{\scalebox{0.78}{0.087}} &{\scalebox{0.78}{0.101}} &{\scalebox{0.78}{0.416}} & \scalebox{0.78}{-} & \scalebox{0.78}{-} &\scalebox{0.78}{0.095} &\scalebox{0.78}{0.088} &\scalebox{0.78}{0.088} &\scalebox{0.78}{0.084} &\scalebox{0.78}{0.093} &\scalebox{0.78}{0.117}  &\scalebox{0.78}{0.100} &\scalebox{0.78}{0.156} \\
    \midrule

    \multirow{4}{*}{\rotatebox{90}{\scalebox{0.95}{Taxi}}}
    &  \scalebox{0.78}{48} & \scalebox{0.78}{\textbf{0.166}} & \scalebox{0.78}{\textbf{0.264}} & \scalebox{0.78}{0.274} & \scalebox{0.78}{0.363} & {{\scalebox{0.78}{\underline{0.217}}}} & \scalebox{0.78}{\underline{0.327}} & \scalebox{0.78}{0.243} & \scalebox{0.78}{0.330} &\scalebox{0.78}{0.272} & \scalebox{0.78}{0.391}  & {\scalebox{0.78}{0.234}} & {\scalebox{0.78}{0.338}} &{\scalebox{0.78}{0.246}} &{\scalebox{0.78}{0.617}} & \scalebox{0.78}{-} & \scalebox{0.78}{-}  &\scalebox{0.78}{0.264} &\scalebox{0.78}{0.348} &\scalebox{0.78}{0.236} &\scalebox{0.78}{0.345} &\scalebox{0.78}{0.259} &\scalebox{0.78}{0.368} 
    &\scalebox{0.78}{0.276} &\scalebox{0.78}{0.388} \\
    & \scalebox{0.78}{96} & \scalebox{0.78}{\textbf{0.356}} & \scalebox{0.78}{\textbf{0.513 }} & \scalebox{0.78}{0.473} & \scalebox{0.78}{0.577} &\scalebox{0.78}{\underline{0.379}} & \scalebox{0.78}{\underline{0.528}} & \scalebox{0.78}{0.469} & \scalebox{0.78}{0.534} & \scalebox{0.78}{0.491} & \scalebox{0.78}{0.590} & {\scalebox{0.78}{0.371}} & {\scalebox{0.78}{0.542}} &\scalebox{0.78}{0.481} &\scalebox{0.78}{0.849} & \scalebox{0.78}{-} & \scalebox{0.78}{-} &\scalebox{0.78}{0.488} & \scalebox{0.78}{0.571} &{\scalebox{0.78}{0.464}} &{\scalebox{0.78}{0.563}} &\scalebox{0.78}{0.476} &\scalebox{0.78}{0.607} &\scalebox{0.78}{0.617} &\scalebox{0.78}{0.625} \\ 
    & \scalebox{0.78}{144} & \scalebox{0.78}{\textbf{0.465}} & \scalebox{0.78}{\textbf{0.538}} & \scalebox{0.78}{0.536} & \scalebox{0.78}{0.724} & \scalebox{0.78}{\underline{0.485}} & \scalebox{0.78}{\underline{0.703}}& \scalebox{0.78}{0.517} & \scalebox{0.78}{0.706} & \scalebox{0.78}{0.532} & \scalebox{0.78}{0.915}  & {\scalebox{0.78}{0.483}} & {\scalebox{0.78}{0.712}}  & \scalebox{0.78}{0.527} &\scalebox{0.78}{1.124} & \scalebox{0.78}{-} & \scalebox{0.78}{-} & \scalebox{0.78}{0.515} &\scalebox{0.78}{0.717} &\scalebox{0.78}{0.522} &\scalebox{0.78}{0.726} &\scalebox{0.78}{0.559} &\scalebox{0.78}{0.774} &{\scalebox{0.78}{0.664}} &\scalebox{0.78}{0.815}\\ 
    \cmidrule(lr){2-26}
    & \scalebox{0.78}{Avg} & \scalebox{0.78}{\textbf{0.329}} & \scalebox{0.78}{\textbf{0.438}} & \scalebox{0.78}{0.428} & \scalebox{0.78}{0.555} & \scalebox{0.78}{\underline{0.360}} & \scalebox{0.78}{\underline{0.519}} & \scalebox{0.78}{0.410} & \scalebox{0.78}{0.523} & \scalebox{0.78}{0.432} & \scalebox{0.78}{0.632}  &{\scalebox{0.78}{0.363}} &{\scalebox{0.78}{0.531}} &\scalebox{0.78}{0.418} &\scalebox{0.78}{0.863} & \scalebox{0.78}{-} & \scalebox{0.78}{-} &\scalebox{0.78}{{0.422}} &\scalebox{0.78}{{0.545}} &\scalebox{0.78}{0.407} &\scalebox{0.78}{0.545} &\scalebox{0.78}{0.431} &\scalebox{0.78}{0.583}
    &\scalebox{0.78}{0.519} &\scalebox{0.78}{0.609} \\
    \midrule
    
    \multirow{3}{*}{\rotatebox{90}{\scalebox{0.80}{Wikipedia}}} 
    &  \scalebox{0.78}{48} & \scalebox{0.78}{0.073} & \scalebox{0.78}{0.095} & \scalebox{0.78}{\textbf{0.063}} & \scalebox{0.78}{\textbf{0.086}} & \scalebox{0.78}{\underline{0.066}} & \scalebox{0.78}{0.093} & \scalebox{0.78}{0.074} & \scalebox{0.78}{\underline{0.090}} & \scalebox{0.78}{0.091} & \scalebox{0.78}{0.142} &\scalebox{0.78}{0.077} &\scalebox{0.78}{0.103} &\scalebox{0.78}{0.112} 
    &\scalebox{0.78}{1.625} &\scalebox{0.78}{-} 
    &\scalebox{0.78}{-} & \scalebox{0.78}{0.081} & \scalebox{0.78}{0.102} &\scalebox{0.78}{0.084} &\scalebox{0.78}{0.111 } &{\scalebox{0.78}{0.083}} &{\scalebox{0.78}{0.109}} &\scalebox{0.78}{0.092} &\scalebox{0.78}{0.107}  \\ 
    & \scalebox{0.78}{96} & \scalebox{0.78}{\textbf{0.074}} & \scalebox{0.78}{\textbf{0.126}} & \scalebox{0.78}{0.086} & \scalebox{0.78}{0.153} & \scalebox{0.78}{\underline{0.080}} & \scalebox{0.78}{\underline{0.137}} & \scalebox{0.78}{0.086} & \scalebox{0.78}{0.143} & \scalebox{0.78}{0.116} & \scalebox{0.78}{0.191} &{\scalebox{0.78}{0.093}} &\scalebox{0.78}{0.146} &\scalebox{0.78}{0.187} &{\scalebox{0.78}{2.234}} & \scalebox{0.78}{-} & \scalebox{0.78}{-} &\scalebox{0.78}{0.119} &\scalebox{0.78}{0.194} &\scalebox{0.78}{0.107} &\scalebox{0.78}{0.148} &\scalebox{0.78}{0.105} &\scalebox{0.78}{0.163} 
    &\scalebox{0.78}{0.131} &\scalebox{0.78}{0.160} \\
    \cmidrule(lr){2-26}
    & \scalebox{0.78}{Avg} & \scalebox{0.78}{\textbf{0.073}} & \scalebox{0.78}{\textbf{0.110}} & \scalebox{0.78}{0.074} & \scalebox{0.78}{0.120} & \scalebox{0.78}{\textbf{0.073}} & \scalebox{0.78}{\underline{0.115}} & \scalebox{0.78}{0.080} & \scalebox{0.78}{0.116} & \scalebox{0.78}{0.104} & \scalebox{0.78}{0.167} &\scalebox{0.78}{0.085} &\scalebox{0.78}{0.125} &\scalebox{0.78}{0.150} &\scalebox{0.78}{1.929} & \scalebox{0.78}{-} & \scalebox{0.78}{-} &\scalebox{0.78}{0.100} &\scalebox{0.78}{0.148} &{\scalebox{0.78}{0.095}} &{\scalebox{0.78}{0.130}} &\scalebox{0.78}{0.094} &\scalebox{0.78}{0.136} &\scalebox{0.78}{0.111} &\scalebox{0.78}{0.135} \\
    \midrule
    \midrule
     \multicolumn{2}{c|}{\scalebox{0.78}{{$1^{\text{st}}$ Count}}} & \scalebox{0.78}{\textbf{16}} & \scalebox{0.78}{\textbf{16}} & \scalebox{0.78}{1} & \scalebox{0.78}{1} & \scalebox{0.78}{\underline{2}} & \scalebox{0.78}{1} & \scalebox{0.78}{1} & \scalebox{0.78}{1} & \scalebox{0.78}{0} & \scalebox{0.78}{0} & \scalebox{0.78}{0} & \scalebox{0.78}{0} & \scalebox{0.78}{0} & \scalebox{0.78}{0} & \scalebox{0.78}{0} & \scalebox{0.78}{0} & \scalebox{0.78}{0} & \scalebox{0.78}{0} & \scalebox{0.78}{0} & \scalebox{0.78}{0} & \scalebox{0.78}{0} & \scalebox{0.78}{0} & \scalebox{0.78}{0} & \scalebox{0.78}{0}\\
    \bottomrule
  \end{tabular}
    \end{small}
  \end{threeparttable}
}
  \caption{Model performance comparisons on the test set $\hat{\operatorname{CRPS}}$:$\operatorname{CRPS}_{\operatorname{sum}}$, $\hat{\text{NRMSE}}$:$\text{NRMSE}_{sum}$ (lower is better) show baselines and our HDT model. – marks out-of-memory failures. Trans-MAF stands for Transformer-MAF. The \underline{underlined} ones as the second best.}\label{tab:full_baseline_results}
\end{table*}

\subsection{Main results}
\textbf{Probabilistic Forecasting Performance}. As shown in Table ~\ref{tab:full_baseline_results}, HDT achieves consistent state-of-the-art performance in most of  benchmarks, covering three prediction settings, large span of dimensions and more showcases are shown in Supplementary due to the page limitation. Especially, HDT achieves a large performance gain over recent popular discrete method VQ-TR, such as average \textbf{25.9}\% $\text{CRPM}_{sum}$ improvement on Traffic, \textbf{23.1}\% $\text{CRPM}_{sum}$ improvement on Taxi. Also, we observe that HDT outperforms some diffusion-based methods TimeDiff, TSDiff and  marginal improvement against recent strong baseline MG\_TSD that is more obvious in the case of high-dimensional and non-stationary datasets, such as average \textbf{13.3}\% improvement on Traffic and \textbf{10.2}\% on Taxi. This implies that the trends of target may introduce more future information gains into our forecasting model. \\
\textbf{Deterministic Forecasting Performance.} In our experiments, we observe that some models exhibit higher values for $\operatorname{CRPS}_{\operatorname{sum}}$, yet lack true predictive accuracy. Therefore, we report the $\text{NRMSE}_{sum}$ for deterministic estimation, which is shown in Table ~\ref{tab:full_baseline_results}. We found that HDT achieves the best results cross all datasets, especially in the Traffic and Taxi datasets, we achieve average improvements of \textbf{15.3}\% and \textbf{15.6}\% $\text{NRMSE}_{sum}$ comparing to the strong baseline MG\_TSD. It is worth noting that  $\operatorname{D}^{3}$VAE and TimeDiff showe significant deviations in point evaluations, but TSDiff with self-guidance demonstrate competitive performance, implying the effectiveness of self-guided strategy. The time and space efficiencies of HDT are shown in Appendix D, due to limited space.
\begin{table*}[t]
\centering
\resizebox{\linewidth}{!}{ 
    \begin{tabular}{@{}|ccccc|cccc|cccc|@{}} 
    \toprule
    \multicolumn{1}{|l|}{Datasets} & \multicolumn{4}{c|}{Electricity} & \multicolumn{4}{c|}{Traffic} &
    \multicolumn{4}{c|}{Taxi}\\ 
    \cmidrule(lr){1-1} \cmidrule(lr){2-5}
    \cmidrule(lr){6-9}
    \cmidrule(lr){10-13}
    \multicolumn{1}{|l|}{Lengths} & \multicolumn{2}{c}{48} & \multicolumn{2}{c|}{96} & \multicolumn{2}{c}{48} & \multicolumn{2}{c|}{96} & \multicolumn{2}{c}{48} & \multicolumn{2}{c|}{96}\\ 
    \cmidrule(lr){1-1} \cmidrule(lr){2-3} \cmidrule(lr){4-5} \cmidrule(lr){6-7} \cmidrule(lr){8-9} \cmidrule(lr){10-11} \cmidrule(lr){12-13}
    \multicolumn{1}{|c|}{Metrics} & $\text{CRPS}_{sum}$ & $\text{NRMSE}_{sum}$ & $\text{CRPS}_{sum}$ & $\text{NRMSE}_{sum}$ & $\text{CRPS}_{sum}$ & $\text{NRMSE}_{sum}$ & $\text{CRPS}_{sum}$ & $\text{NRMSE}_{sum}$ & $\text{CRPS}_{sum}$ & $\text{NRMSE}_{sum}$ & $\text{CRPS}_{sum}$ & $\text{NRMSE}_{sum}$ \\
    \midrule
    \multicolumn{1}{|c|}{C-Transformer} & 0.327(.004) & 0.532(.018) & 0.212(.007)& 0.228(.014) & 0,467(.011) & 0.845(.009) & 1.004(.005) & 1.150(.012) & 0.861(.006) & 1.121(.013) & 0.977(.008) & 1.118(.011)  \\
    \multicolumn{1}{|c|}{HDT} & 0.025(.002) &  0.030(.002) &  0.028(.001) & 0.032(.003) & 0.034(.001) & 0.060(.004) & 0.037(.003) & 0.063(.005) & 0.166(.005) & 0.264(.003) & 0.356(.002) & 0.513(.007)  \\
    \bottomrule
    \end{tabular}

}\caption{Performance of HDT with Continuous Transformer structure C-Trasformer, which does not include the quantization layer in the stage 1 and replace the discrete token sequences of HDT with continuous representation from stage 1.}
    \label{C-Transformer}
\end{table*}

 \begin{figure*}[!t]
  \centering
  \includegraphics[width=1.0\linewidth]{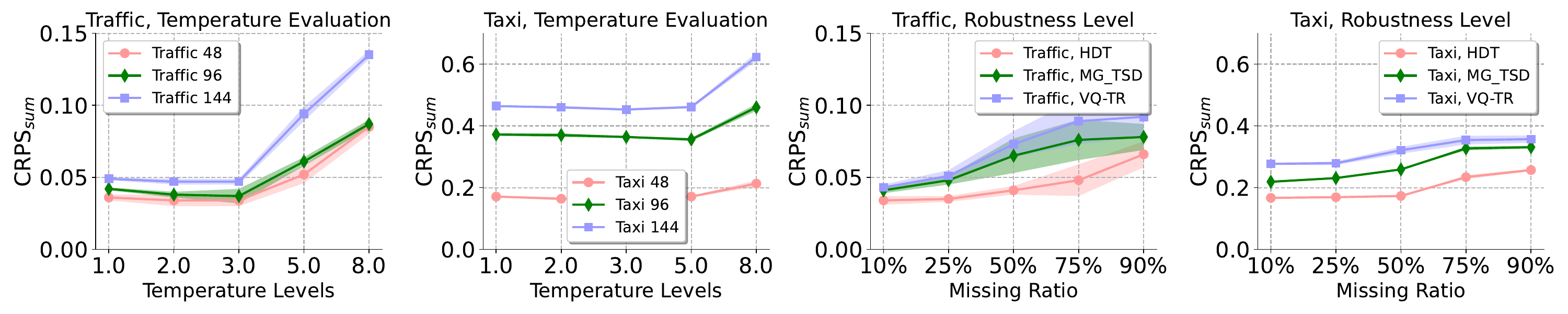}
  \caption{Performance of HDT with different temperature levels of different prediction lengths in Traffic and Taxi datasets. The comparison results against MG\_TSD and VQ-TR with HDT on different levels of missing rate.}
  \label{Temperature}
\end{figure*}
\begin{figure*}[ht]
  \centering
\includegraphics[width=1.0\linewidth]{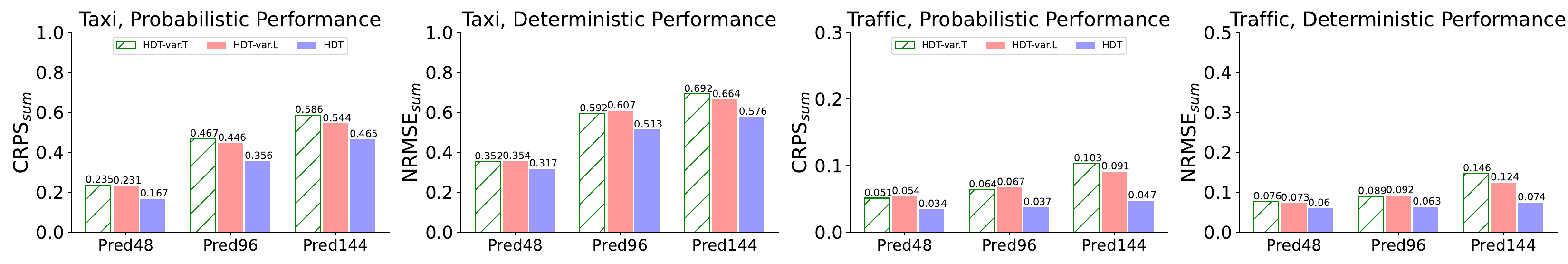}
  \caption{Probabilistic and deterministic performance of HDT and HDT-variants on different prediction length and datasets. HDT-var.T is the same structure with HDT without the self-conditions in stage 2. HDT-var.L replaces the Transformer with LSTM in stage 2 and without self-conditions.}
  \label{discrete_conti}
\end{figure*} 
\begin{table*}[!t]
\centering
\resizebox{\linewidth}{!}{ 
    \begin{tabular}{@{}|ccccccc|cccccc|@{}} 
    \toprule
    \multicolumn{1}{|l|}{Datasets} & \multicolumn{6}{c|}{Traffic} & \multicolumn{6}{c|}{Taxi} \\ 
    \cmidrule(lr){1-1} \cmidrule(lr){2-7} \cmidrule(lr){8-13}
    \multicolumn{1}{|l|}{Lengths} & \multicolumn{2}{c}{48} & \multicolumn{2}{c}{96} & \multicolumn{2}{c|}{144} & \multicolumn{2}{c}{48} & \multicolumn{2}{c}{96} & \multicolumn{2}{c|}{144}\\ 
    \cmidrule(lr){1-1} \cmidrule(lr){2-3} \cmidrule(lr){4-5} \cmidrule(lr){6-7} \cmidrule(lr){8-9} \cmidrule(lr){10-11} \cmidrule(lr){12-13}
    \multicolumn{1}{|c|}{Metrics} & $\text{CRPS}_{sum}$ & $\text{NRMSE}_{sum}$ & $\text{CRPS}_{sum}$ & $\text{NRMSE}_{sum}$ & $\text{CRPS}_{sum}$ & $\text{NRMSE}_{sum}$ & $\text{CRPS}_{sum}$ & $\text{NRMSE}_{sum}$ & $\text{CRPS}_{sum}$ & $\text{NRMSE}_{sum}$ & $\text{CRPS}_{sum}$ & $\text{NRMSE}_{sum}$ \\
    \midrule
    \multicolumn{1}{|c|}{HDT-$\text{h}_{c}$} & 0.042(.003) & 0.069(.005) & 0.046(.002) & 0.077(.006) & 0.056(.002) & 0.084(.003) & 0.206(.004) & 0.316(.005) & 0.373(.006) & 0.548(.010) & 0.481(.004) & 0.573(.006)   \\
    \multicolumn{1}{|c|}{HDT-$\text{h}_{d^{*}}$} & 0.038(.002) & 0.065(.008) & 0.039(.004) & 0.067(.004) & 0.050(.004) & \underline{0.078(.003)} & 0.189(.004) & 0.278(.004) & 0.371(.004) & 0.537(.005) & 0.480(.005) & 0.568(.011)  \\
    \multicolumn{1}{|c|}{HDT-$\text{h}_{d}$} & \textbf{0.034(.001)} & \textbf{0.060(.004)} & \underline{0.037(.003)} & \textbf{0.063(.005)} & \underline{0.048(.005)} & 0.079(.006) & \textbf{0.166(.003)} & \textbf{0.264(.005)} & \textbf{0.356(.002)} & \textbf{0.513(.004)} & \underline{0.467(.006)} & \underline{0.540(.007)}  \\
    \multicolumn{1}{|c|}{HDT-$\text{h}_{dc}$} & \underline{0.036(.003)} & \underline{0.062(.006)} & \textbf{0.037(.001)} & \underline{0.064(.002)} & \textbf{0.047(.004)} & \textbf{0.076(.008)} & \underline{0.171(.005)} & \underline{0.266(.003)} & \underline{0.356(.004)} & \underline{0.517(.009)} & \textbf{0.465(.003)} & \textbf{0.537(.008)}  \\
    \bottomrule
    \end{tabular}
}\caption{Performance of HDT with Different Types of contextual conditions.\textbf{Bold} numbers represent the best outcomes and the \underline{underlined} ones as the second best.}
\label{Effect_conds}
\end{table*}
\begin{table*}[!t]
\centering
\resizebox{\linewidth}{!}{ 
    \begin{tabular}{@{}|ccccccc|cccccc|@{}} 
    \toprule
    \multicolumn{1}{|l|}{Datasets} & \multicolumn{6}{c|}{Traffic} & \multicolumn{6}{c|}{Taxi} \\ 
    \cmidrule(lr){1-1} \cmidrule(lr){2-7} \cmidrule(lr){8-13}
    \multicolumn{1}{|l|}{Lengths} & \multicolumn{2}{c}{48} & \multicolumn{2}{c}{96} & \multicolumn{2}{c|}{144} & \multicolumn{2}{c}{48} & \multicolumn{2}{c}{96} & \multicolumn{2}{c|}{144}\\ 
    \cmidrule(lr){1-1} \cmidrule(lr){2-3} \cmidrule(lr){4-5} \cmidrule(lr){6-7} \cmidrule(lr){8-9} \cmidrule(lr){10-11} \cmidrule(lr){12-13}
    \multicolumn{1}{|c|}{Metrics} & $\text{CRPS}_{sum}$ & $\text{NRMSE}_{sum}$ & $\text{CRPS}_{sum}$ & $\text{NRMSE}_{sum}$ & $\text{CRPS}_{sum}$ & $\text{NRMSE}_{sum}$ & $\text{CRPS}_{sum}$ & $\text{NRMSE}_{sum}$ & $\text{CRPS}_{sum}$ & $\text{NRMSE}_{sum}$ & $\text{CRPS}_{sum}$ & $\text{NRMSE}_{sum}$ \\
    \midrule
    \multicolumn{1}{|c|}{2} & \underline{0.036(.003)} & \underline{0.070(.005)} & 0.044(.002) & 0.079(.007) & 0.061(.004) & 0.094(.008) & 0.226(.005) & 0.338(.006) & 0.373(.007) & 0.530(.012) & 0.542(.005) & 0.714(.012)  \\
    \multicolumn{1}{|c|}{3} & 0.036(.005) & 0.073(.003) & \textbf{0.037(.003)} & \textbf{0.063(.005)} & \underline{0.049(.003)} & \underline{0.080(.006)} & \textbf{0.166(.003)} & \textbf{0.264(.005)} & 0.378(.005) & 0.569(.009) & 0.530(.008) & 0.635(.014)  \\
    \multicolumn{1}{|c|}{4} & \textbf{0.034(.001)} & \textbf{0.060(.004)} & 0.039(.004) & 0.067(.006) & \textbf{0.047(.004)} & \textbf{0.076(.008)} & \underline{0.172(.002)} & 0.277(.007) & \textbf{0.356(.002)} & \textbf{0.513(.004)} & \textbf{0.465(.003)} & \textbf{0.537(.008)}  \\
    \multicolumn{1}{|c|}{5} & 0.037(.002) & 0.075(.003) & \underline{0.363(.003)} & \underline{0.524(.005)} & 0.052(.002) & 0.082(.005) & 0.173(.003) & \underline{0.268(.004)} & \underline{0.363(.004)} & \underline{0.526(.007)} & \underline{0.476(.007)} & \underline{0.578(.011)}  \\
    \bottomrule
    \end{tabular}

}\caption{Performance of HDT with Transformer layers under different prediction lengths on Traffic (Stationary) and Taxi (Non-stationary). We report mean\&stdev.results of 3 runs.}
    \label{Effect_Layers}
\end{table*}

\subsection{Ablation studies}
\textbf{Effect of Discrete Representation $z_q$ in Eqn.~(\ref{33333}).} To verify the effectiveness of discrete representations in MTS, we conducted an experiment by bypassing the discretization of the intermediate variable $\hat{z}$ in stage 1, directly inputting it into stage 2 for autoregressive generation via cross-attention with the context encoder. We tested this on three datasets (Electricity, Traffic, Taxi) with two prediction lengths (48 and 96), covering dimensions from 370 to 1214. As shown in Table~\ref{C-Transformer}, the continuous structure (C-Transformer) performed poorly in both probabilistic and deterministic scenarios. We believe that without discretization, $\hat{z}$ acts as an infinitely large codebook, making it difficult for the stage 2 Transformer to fit properly. This highlights the effectiveness of our discrete Transformer structure. \\
\textbf{Effect of Historical Condition $\mathcal{H}^{p}$ in Eqn.~(\ref{selfcond_cond}).} To verify the applicability of discrete representations in multivariate time series, we set four different forms of historical sequences during the second stage of training: (i) HDT-$\text{h}_{c}$: the continuous features from the stage1, not transformed into discrete form; (ii) HDT-$\text{h}_{d}$: transformed into the corresponding discrete form in the stage 1; (iii) HDT-$\text{h}_{d^{*}}$: the discrete features, without entering the Encoder of stage 2 (iv) HDT-$\text{h}_{dc}$: concatenation of discrete and continuous representations from the stage 1. We test on two high-dimensional and distinct types of multivariate time series and the results are shown in Table ~\ref{Effect_conds}, the relatively stable and periodic Traffic, and the Taxi series, which is of higher frequency of fluctuations and more outliers. We observe that in the Traffic and Taxi of all prediction settings, HDT-$\text{h}_{c}$ performs obviously lower than HDT-$\text{h}_{d}$ and HDT-$\text{h}_{d^{*}}$, while HDT-$\text{h}_{dc}$ is competitive. This suggests that within a probabilistic framework, discrete representations, serving as an approximate expression, can be seen as a ``Clustering" result that is more resilient to stochastic changes. By incorporating target trends, HDT can achieve a higher level of deterministic forecasting performance.\\
\textbf{Effect of Discrete Self-Condition $s_{down}$ in Eqn.~(10).} From Figure ~\ref{discrete_conti}, we have: i) For the short-term prediction length (e.g.48) of two datasets, both HDT-var.T and HDT-var.L show marginal differences between HDT, implying the effectiveness of discrete representations. In contrast, these variants show obvious differences between HDT of 96 and 144 settings, which further verifies the merits of our self-conditioned strategy.ii) Discrete features demonstrate stable performance in relatively steady dataset(e.g.Traffic), without significant declines as the forecasting horizon extends. However, in non-stationary dataset(e.g.Taxi), it still exhibits notable performance fluctuations of discrete representations, which implies the effectiveness of our self-condition strategy.\\
\noindent \textbf{Effect of Missing Ratios in Eqn.~(\ref{selfcond_cond}).} To evaluate HDT's robustness, we implemented a timestamp masking strategy, allowing the network to infer representations under incomplete contexts. We randomly masked observations (historical sequences) in the test sets of the Traffic (pred 96) and Taxi (pred 48) datasets at designated missing rates. Figure~\ref{Temperature} illustrates that excluding the target condition from the forecasting model leads to a rapid decline in probabilistic performance as the missing rate increases in two diffusion models. From the Taxi dataset, with the missing rate of historical conditions nearing 100\%, HDT's performance remains largely unaffected, in contrast to the obvious performance degradation observed in the other two history-conditioned diffusion models.\\
\textbf{Effect of Temperature Levels in Inference.} During our experiments, we observed that sampling temperature is a crucial hyperparameter in a probabilistic setting. As shown in Figure~\ref{Temperature}, tests on the Traffic and Taxi datasets revealed significant differences in results with varying temperatures. As for the Traffic dataset, a slightly higher temperature improved probabilistic forecasting performance, while a substantial increase led to model bias. For the Taxi dataset, we found that a moderate temperature is optimal, with no significant change in short-term accuracy at higher temperatures compared to long-term settings. This suggests that HDT can achieve better results by adjusting temperature variations to suit different datasets and forecasting lengths.\\
\textbf{Effect of Number of Layers in Eqn.~(\ref{selfcond_cond}).} To investigate the effect of the self-cond Transformer layers in Eqn. (\ref{selfcond_cond}), we report the $\text{CRPS}_{sum}$ and $\text{NRMSE}_{sum}$ results of our SDT with different number of layers (e.g.2, 3, 4, 5) in Table ~\ref{Effect_Layers}. We observe that in short-term forecasting, a smaller number of layers (e.g., 2, 3) shows competitive results in both datasets. As the forecast length increases, Traffic exhibits superior performance with a moderately increased number of layers, while high-stochastic Taxi excels in deeper Transformer structures. These experimental results were all conducted under the condition that the base Transformer decoder layers in Eqn.~\ref{1212} are fixed at 3.

\section{Conclusion} In this paper, we propose a hierarchical self-conditioned discrete method \textbf{HDT} to enhance high-dimensional multivariate time series (MTS) forecasting. Our novel two-stage vector quantized generative framework maps targets into discrete token representations, capturing target trends for long-term forecasting. To the best of our knowledge, this is the first discrete Transformer architecture applied to high-dimensional, long-term forecasting tasks. Extensive experiments on benchmark datasets demonstrate the effectiveness of our approach. Future research will explore integrating multimodal data into MTS forecasting.

\section{Acknowledgments}
This research is supported by the Joint NTU-WeBank Research Centre on Fintech, Nanyang Technological University, Singapore.

\bibliography{anonymous-submission-latex-2025}

\clearpage
\appendix
\section{Appendix}
In the supplementary, we provide more implementation details, more experimental results, and visualization of test samples of our HDT. We organize our supplementary as follows
\begin{itemize}
\item In Section A, we give the Related Work of HDT, including vector quantization-based frameworks and deep generative model-based MTS two parts.
\item In Section B, we provide more details of used datasets and metrics in our experiment.
\item In Section C, we provide the experiment setup, including the hyperparameters and detailed structures of stage 1, 2 frameworks. 
\item In Section D, we draw the comparison of the memory usage and inference time between HDT and other strong baselines, highlighting the source-efficient and efficiency.
\item In Section E, we show more experimental results wtih obvious non-stationary datasets Hospital \cite{hyndman2008forecasting} and COVID Deaths \cite{dong2020interactive} with different prediction lengths \{24, 48, 96\} to compare the prediction performance  with state-of-the-art generative methods.
\item In Section F, we provide the details of baselines in our main experiment.
\item In Section G, we summarize the limitations and showcase more test samples on seven MTS datasets.
\end{itemize}

\section{A. Related work} 
\label{Appendix_related}
\subsection{Vector quantization-based frameworks} Unlike many deep learning methods directly focusing on the continuous data domains, Vector Quantization-based frameworks map complex continuous domains into finite discrete domains. VQVAE \cite{van2017neural, razavi2019generating} decomposes the image generation process into two parts: initially, it trains a vector quantized autoencoder aimed at image reconstruction, transforming images into a compressed sequence of discrete tokens. Then the second stage learns an autoregressive model, e.g., PixelSNAIL \cite{chen2018pixelsnail}, to model the underlying distribution of token sequences. Driven by the effectiveness of VQVAE and progress in sequence modeling, many approaches follow the two-stage paradigm. DALL-E \cite{ramesh2021zero} improves token prediction in the second stage by using Transformers, resulting in a strong text-to-image synthesis model. VQGAN \cite{esser2021taming, zheng2022movq, yu2021vector} employs adversarial loss during its first stage, training a more efficient autoencoder, which allows for the synthesis of images with greater details.

\subsection{Deep generative model-based MTS}

To improve the reliability and performance of high-dimensional MTS, instead of modeling the raw data, there exist works inferring the underlying distribution of the time series data with deep generative models \cite{yoon2019time, brophy2023generative}.  Normalizing flow \cite{papamakarios2017masked, dinh2016density} based MTS framework, e.g., MAF \cite{rasul2020multivariate} explicitly models multivariate time series and their temporal dynamics by employing a normalizing flow for probabilistic forecasting. Variational Autoencoder-based models, e.g., Timevae \cite{desai2021timevae} a novel architecture with interpretability, can encode domain knowledge, and reduce training times. 

Existing diffusion-based \cite{ho2020denoising} MTS forecasting models can be roughly divided into two categories. The first one is autoregressive, e.g., TimeGrad \cite{rasul2021autoregressive} operates by sequentially generating future predictions over time. Nonetheless, its ability to forecast over long ranges is constrained by the accumulation of errors and a sluggish inference speed. The other is the non-autoregressive diffusion model, such as CSDI \cite{tashiro2021csdi}, LDT \cite{feng2024latent}, SSSD \cite{alcaraz2022diffusion}, $\operatorname{D}^{3}$VAE \cite{li2022generative}, TSDiff \cite{kollovieh2023predict}, TimeDiff \cite{shen2023non}, MG\_TSD \cite{fan2024mg} and TMDM \cite{litransformer}. These models perform conditioning and unconditioning strategies to train the denoising networks and introduce some guidance strategies to predict the denoising objective more accurately. 

\section{B. Dataset and Metric Details} 
\label{Appendix_dataset}
\textbf{Dataset.} We summarize the dataset details of our MTS long-term forecasting. As shown in Table ~\ref{tab:dataset}, Solar is the hourly photo-voltaic production of 137 stations in Alabama State; Electricity is the hourly time series of the electricity consumption of 370 customers; Traffic is the hourly occupancy rate, between 0 and 1, of 963 San Francisco car lanes; Taxi is the spatio-temporal half hourly traffic time series of New York taxi rides taken at 1214 locations; Wikipedia \cite{gasthaus2019probabilistic} is the daily page views of 2000 Wikipedia pages. Among them, the Solar shows certain periodic patterns, whereas the others predominantly display non-stationary characteristics.
\begin{table}[htbp]
  \centering
\resizebox{1.0\linewidth}{!}{\begin{tabular}{|l|cccccc|}
\hline
  \small DATASET & \small Dimension & \small Domain & \small Freq & \small Total Time Steps & \small Context Length & \small Pred Length  \\ \hline
\small Solar  & \small 137  & \small $\mathbb{R}^{+}$ & \small Hourly & \small 7,009 & \small 96 & \small \{48, 96, 144\} \\ 
\small COVID Deaths  & \small 266  & \small $\mathbb{R}^{+}$ & \small Daily & \small 212 & \small 96 & \small \{48, 96\} \\ 
\small Electricity  & \small 370 & \small $\mathbb{R}^{+}$ & \small Hourly & \small 5,790 & \small 96&  \small \{48, 96, 144\}  \\  
\small Hospital  & \small 767 & \small $\mathbb{R}^{+}$ & \small Monthly & \small 84 & \small 24 &  \small \{24, 48\}  \\ 
\small Traffic   & \small 963 & \small (0,1) & \small Hourly & \small 10,413 & \small 96& \small \{48, 96, 144\} \\  
\small Taxi    & \small 1214  & \small $\mathbb{N}$ & \small 30-Min & \small 1,488 & \small 96 & \small \{48, 96, 144\} \\ 
\small Wikipedia   & \small 2000  & \small $\mathbb{N}$ & \small Daily & \small 792 & \small 96 & \small \{48, 96\}\\ \hline
\end{tabular}
} \caption{Properties of the datasets in experiments}
  \label{tab:dataset}
\end{table} \\

\begin{table*}[!ht]
\centering
\resizebox{\linewidth}{!}{ 
    \begin{tabular}{@{}|ccccc|cccc|cccc|@{}} 
    \toprule
    \multicolumn{1}{|l|}{Datasets} & \multicolumn{4}{c|}{Electricity} & \multicolumn{4}{c|}{Traffic} &
    \multicolumn{4}{c|}{Taxi}\\ 
    \cmidrule(lr){1-1} \cmidrule(lr){2-5}
    \cmidrule(lr){6-9}
    \cmidrule(lr){10-13}
    \multicolumn{1}{|l|}{Lengths} & \multicolumn{2}{c}{48} & \multicolumn{2}{c|}{96} & \multicolumn{2}{c}{48} & \multicolumn{2}{c|}{96} & \multicolumn{2}{c}{48} & \multicolumn{2}{c|}{96}\\ 
    \cmidrule(lr){1-1} \cmidrule(lr){2-3} \cmidrule(lr){4-5} \cmidrule(lr){6-7} \cmidrule(lr){8-9} \cmidrule(lr){10-11} \cmidrule(lr){12-13}
    \multicolumn{1}{|c|}{Metrics} & $\text{QICE} \downarrow$ & $\text{PICP} \uparrow $ & $\text{QICE} \downarrow$ & $\text{PICP} \uparrow$ & $\text{QICE} \downarrow$ & $\text{PICP} \uparrow$ & $\text{QICE} \downarrow$ & $\text{PICP} \uparrow$ & $\text{QICE} \downarrow$  & $\text{PICP} \uparrow$ & $\text{QICE} \downarrow$ & $\text{PICP} \uparrow$ \\
    \midrule
    \multicolumn{1}{|c|}{TimeGrad} & 10.17 & 80.16 & 12.29 & 77.23 & 12.36 & 88.24 & 14.73 & 86.39 & 6.47 & 41.76 & 8.76 & 38.23 \\
    \multicolumn{1}{|c|}{HDT} & 7.74 & 83.72 & 7.96 & 82.94 & 4.72 & 95.27 & 6.39 & 96.23 & 4.36 & 54.03 & 6.63 & 49.24  \\
    \bottomrule
    \end{tabular}

}\caption{Probabilistic forecasting performance of HDT with TimeGrad on the QICE and PICP, whih are popular metrics for  probabilistic multivariate time series forecasting models}
    \label{QICE}
\end{table*}

\noindent \textbf{Metric.} CRPS measures the compatibility of a cumulative distribution function $P$ with an observation $x$ as: $\operatorname{CRPS}(\mathcal{F}, x)=\int_{\mathbb{R}}(P(y)-\mathbb{I}\{x \leq y\})^{2} d y,$ where $\mathbb{I}\{x \leq y\}$is the indicator function which is one if $x \leq y$ and zero otherwise. The empirical CDF of $P$, i.e., $\hat{P}(y)=\frac{1}{N} \sum_{i=1}^{N} \mathbb{I}\left\{X_{i} \leq y\right\}$ with n samples $X_{i} \sim P$ as the approximation of the predictive CDF. It utilizes $N$ samples to estimate the empirical CDF and take the CRPS-sum in the multivariate case.
\begin{equation}
\vspace{-0.1in}
\operatorname{CRPS_{\text {sum }}}=\mathbb{E}_{t}\left[\operatorname{CRPS}\left(\widehat{P}_{\text {sum }}(t), \sum_{i} x_{i}^{t}\right)\right].
\end{equation} \\
The Normalized Root Mean Squared Error (NRMSE) is a standardized version of the Root Mean Squared Error (RMSE) that accounts for the scale of the target values. The formula for NRMSE is given below:
\begin{equation}
\text{NRMSE}=\sqrt{\frac{1}{T} \sum_{t=1}^T\left(\frac{y_{t}-\hat{y_{t}}}{y_{max}-y_{min}}\right)^2}, 
\end{equation}
where $\hat{y_{t}}$ represents the predicted target, and $y_{t}$ represents the true target. $y_{max}$ and $y_{min}$ are the minimum and maximum of the measured target values, respectively. The NRMSE quantifies the average squared discrepancy between the predictions and actual observations, normalized by the range of the target values. A lower NRMSE indicates higher predictive accuracy.

\section{C. Detailed Experiment Setup and Architectures}
\label{Appendix_Exe}
\begin{table*}[htbp]
\newcommand{\tabincell}[2]{\begin{tabular}{@{}#1@{}}#2\end{tabular}}
 \begin{center}
 \begin{threeparttable}
    \resizebox{1.0\linewidth}{!}{
 	\begin{tabular}{|l|ccccc|ccccc|}
 	\multicolumn{1}{c}{} & \multicolumn{5}{c}{Stage 1} & \multicolumn{5}{c}{Stage 2}  \\
 	\toprule
 	 Dataset & Codebook Size & Codebook dim & Hidden dim & Enc/Dec Layers & Trasformer Layers & Hidden dim &History Encoder & Base Layers & Self-cond Layers & Temperature\\
 	\midrule
        Solar & 128 & \{64, 128, 128\} & \{64, 128, 128\} & 3 & 2 & \{64, 128\} & 2 & 3 & \{3, 4, 5\} & [1.0, 1.5, 2.0, 3.0, 6.0] \\
        Electricity & 128 & 128 & 128 & 3 & 2 & 128 & 2 & 3 & \{3, 4, 5\} &  [1.0, 2.0, 3.0, 5.0, 8.0] \\
        Traffic & 128 & 256 & 256 & 3 & 2 & 256 & 2 & 3 & \{3, 4, 5\} & [1.0, 2.0, 3.0, 5.0, 8.0]  \\
        Taxi & 256 & 256 & 256 & 3 & 3 & 256 & 2 & 3 & \{3, 4, 5\} & [1.0, 2.0, 3.0, 5.0, 8.0] \\
        Wikipedia & 256 & 512 & 512 & 3 & 2  & 512 & 2 & 3 & \{3, 4, 5\} & [1.0, 2.0, 3.0, 5.0, 8.0]\\
    \bottomrule
	\end{tabular}
	}
	 \end{threeparttable}
	 \end{center}
    \caption{Detailed hyperparameters of stages 1 and 2.}
  \label{hyperparameters}
\end{table*}

In this section, we summarize the detailed experiment setup of our HDT. Table ~\ref{hyperparameters} shows the hyperparameters of our overall structure in stages 1 and 2. \{*\} represents the hyperparameters used in our experiments. Table 8, 9 and 10 show the detailed modules of our HDT, among these components, Conv1d refers to the 1-d convolution operation, while the Self-Attn Block and Cross-Attn Block represent the standard multi-head self-attention and cross-attention of TransformerDecoderLayer, respectively.
\begin{table*}[bhp]
    \label{tab:Tran-Enc}
\newcommand{\tabincell}[2]{\begin{tabular}{@{}#1@{}}#2\end{tabular}}
 \begin{center}
 \begin{threeparttable}
    \resizebox{0.80\linewidth}{!}{
 	\begin{tabular}{|c|c|c|}
 	\multicolumn{1}{c}{Layer} & \multicolumn{1}{c}{Function} & \multicolumn{1}{c}{Descriptions} \\
        \toprule
 	 1 & Convolution & input channel=H, output channel=D, kernel size=4, stride=2, padding=1\\
          2   & ReLU & nn.ReLU() \\
       3 & Dropout & nn.Dropout(p=0.1) \\
       4 & LayerNorm & nn.LayerNorm() \\
       5 & Convolution & input channel=D, output channel=D, kernel size=3, stride=1, padding=1\\
       6   & ReLU & nn.ReLU() \\
       7 & Dropout & nn.Dropout(p=0.1) \\
       8 & LayerNorm & nn.LayerNorm() \\
       9 & Convolution & input channel=D, output channel=D, kernel size=3, stride=1, padding=1\\
       10   & Tanh & nn.Tanh() \\
    \bottomrule
	\end{tabular}
	}
	 \end{threeparttable}
	 \end{center}
    \caption{
    The detailed architecture of the Conv-Enc.
    }
\end{table*}
\begin{table*}[ht]
    \label{tab:Deconv-Dec}
\newcommand{\tabincell}[2]{\begin{tabular}{@{}#1@{}}#2\end{tabular}}
 \begin{center}
 \begin{threeparttable}
    \resizebox{0.80\linewidth}{!}{
 	\begin{tabular}{|c|c|c|}
 	\multicolumn{1}{c}{Layer} & \multicolumn{1}{c}{Function} & \multicolumn{1}{c}{Descriptions} \\
        \toprule
 	 1 & DeConvolution & input channel=D, output channel=D, kernel size=3, stride=1, padding=1\\
          2   & ReLU & nn.ReLU() \\
          3 & Dropout & nn.Dropout(p=0.1) \\
          4 & LayerNorm & nn.LayerNorm() \\
          5 & DeConvolution & input channel=D, output channel=D, kernel size=3, stride=1, padding=1\\
          6   & ReLU & nn.ReLU() \\
          7 & Dropout & nn.Dropout(p=0.1) \\
          8 & LayerNorm & nn.LayerNorm() \\
          9 & DeConvolution & input channel=D, output channel=H, kernel size=4, stride=2, padding=1\\
    \bottomrule
	\end{tabular}
	}
	 \end{threeparttable}
	 \end{center}
    \caption{The detailed architecture of the DeConv-Dec.
    }
\end{table*}

\begin{table*}[htbp]
    \label{tab:self-transformer}
\newcommand{\tabincell}[2]{\begin{tabular}{@{}#1@{}}#2\end{tabular}}
 \begin{center}
 \begin{threeparttable}
    \resizebox{0.80\linewidth}{!}{
 	\begin{tabular}{|c|c|c|}
 	\multicolumn{1}{c}{Layer} & \multicolumn{1}{c}{Function} & \multicolumn{1}{c}{Descriptions} \\
        \toprule
 	 1 & Layernorm & nn.LayerNorm()\\
          2 & Self-attention & Attention(q=x, k=x, v=x) \\
          3 & Cross-attention & Attention(q=x, k=history, v=history) \\
          4 & Self-condition attention & Attention(q=x, k=downsampled x, v=downsampled x) \\
          5 & Layernorm & nn.LayerNorm() \\
          6 & MLP & nn.Linear() \\
          7 & ReLU & nn.ReLU()\\
          8 & MLP & nn.Linear() \\
    \bottomrule
	\end{tabular}
	}
	 \end{threeparttable}
	 \end{center}
        \caption{
    The detailed architecture of the Self-Transformer block.
    }
\end{table*}
\section{D. Memory Usage and Model Efficiency}  
\label{Appendix_memory}
\begin{figure*}[ht]
  \centering
  \includegraphics[width=0.8\linewidth]{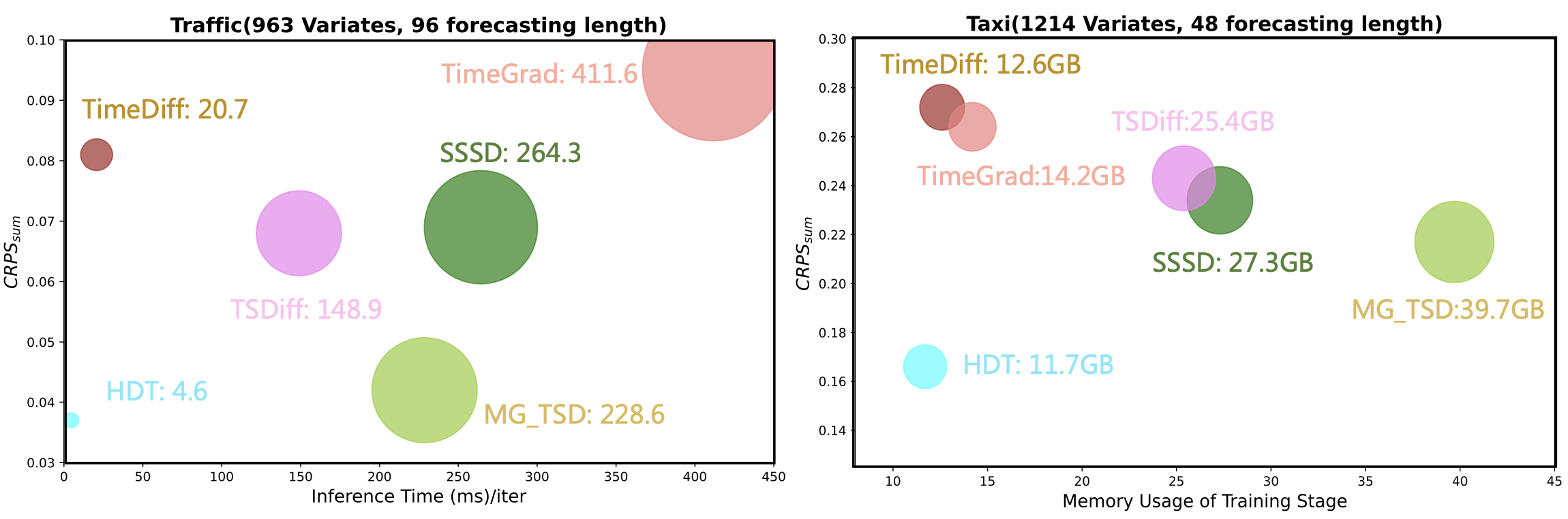}
  \caption{Model memory usage and time efficiency comparison under input-96-predict-{48, 96} of Traffic and Taxi, respectively.}
  \label{Memory}
\end{figure*}

We comprehensively compare the performance of inference time and memory usage of the following models: TimeGrad, SSSD, TimeDiff, TSDiff and MG\_TSD with our efficient discrete framework. The results are recorded with the official model configuration and the same samples numbers=100. In Figure~\ref{Memory}, we compare the efficiency under two representative datasets (963 variates in Traffic and 1214 in Taxi) with different forecasting length (Traffic:96, Taxi:48) and same 96 time steps for lookback. 

From Figure~\ref{Memory}, we observe that HDT demonstrates a significant advantage in memory usage and inference time compared to diffusion models on high-dimensional multivariate time series. It's evident that diffusion-based models tend to increase diffusion steps significantly to enhance prediction performance, especially in high-dimensional data. For example, MG\_TSD introduces the concept of multiple granularities, which not only suffers from the inherent limitations of autoregressive structures but also adds additional inference results from different granularities, thereby further slowing down the inference speed and increasing model size. Non-autoregressive forms like TimeDiff sacrifice some inference accuracy to expedite inference speed. HDT, on the other hand, leverages a compressed discrete structure without relying on a large diffusion framework, which enhances prediction accuracy by utilizing the target itself while ensuring the model is both source-efficient and time-efficient.

\begin{table*}[htbp]
\centering
\resizebox{\linewidth}{!}{ 
    \begin{tabular}{@{}|ccccc|cccc|@{}} 
    \toprule
    \multicolumn{1}{|l|}{Datasets} & \multicolumn{4}{c|}{Hospital} & \multicolumn{4}{c|}{Covid Deaths} \\ 
    \cmidrule(lr){1-1} \cmidrule(lr){2-5} \cmidrule(lr){6-9}
    \multicolumn{1}{|l|}{Lengths} & \multicolumn{2}{c}{inputs:24-forecast:24} & \multicolumn{2}{c|}{inputs:24-forecast:48} & \multicolumn{2}{c}{inputs:48-forecast:48} & \multicolumn{2}{c|}{inputs:96-forecast:96} \\ 
    \cmidrule(lr){1-1} \cmidrule(lr){2-3} \cmidrule(lr){4-5} \cmidrule(lr){6-7} \cmidrule(lr){8-9}
    \multicolumn{1}{|c|}{Metrics} & $\text{CRPS}_{sum}$ & $\text{NRMSE}_{sum}$ & $\text{CRPS}_{sum}$ & $\text{NRMSE}_{sum}$ & $\text{CRPS}_{sum}$ & $\text{NRMSE}_{sum}$ & $\text{CRPS}_{sum}$ & $\text{NRMSE}_{sum}$\\
    \midrule
    \multicolumn{1}{|c|}{TSDiff} & 0.059(.002) & 0.085(.001) & 0.089(.001) & 0.142(.000) & 0.167(.012) & 0.196(.024) & 0.224(.016) & 0.248(.014)  \\
    \multicolumn{1}{|c|}{MG\_TSD} & 0.051(.001) & 0.074(.000) & 0.094(.001) & 0.158(.001) & 0.154(.008) & 0.172(.014) & 0.207(.013) & 0.231(.009)  \\
    \multicolumn{1}{|c|}{HDT-var.T} & 0.037(.003) & 0.047(.002) & 0.072(.003) & 0.091(.001) & 0.134(.008) & 0.169(.011) & 0.211(.008) & 0.236(.014) \\
    \multicolumn{1}{|c|}{HDT \textbf{(ours)}} & \textbf{0.035(.001)} & \textbf{0.043(.001)} & \textbf{0.057(.002)} & \textbf{0.066(.001)} & \textbf{0.127(.006)} & \textbf{0.165(.008)} & \textbf{0.154(.011)} & \textbf{0.178(.007)} \\
    \multicolumn{1}{|c|}{improvement $\textcolor{red}{\uparrow}$} & 31\% $\textcolor{red}{\uparrow}$ & 41\% $\textcolor{red}{\uparrow}$ & 35.2\% $\textcolor{red}{\uparrow}$ & 53.7\% $\textcolor{red}{\uparrow}$ & 17.2\%$\textcolor{red}{\uparrow}$ & 4.3\% $\textcolor{red}{\uparrow}$ & 25.6\% $\textcolor{red}{\uparrow}$ & 22.9\% $\textcolor{red}{\uparrow}$ \\
    
    \bottomrule
    \end{tabular}
    } 
    \caption{Performance of HDT with TSDiff and MG\_TSD on two non-stationary datasets, Hospital and Covid Deaths.\textbf{Bold} numbers represent the best outcomes and the \underline{underlined} ones as the second best.}
\label{390}
\end{table*}

\section{E. More Experiment Results}
\label{experiment}
To validate the advantages of HDT in high-dimensional, non-stationary datasets, we added COVID Deaths and Hospital, as shown in Table ~\ref{390}. From Table ~\ref{390}, we demonstrate significant performance improvements in the complex Hospital dataset. Notably, HDT-var.T, a discrete Transformer without a self-condition strategy, outperforms diffusion-based methods in short-term forecasting settings. However, it struggles to adapt to increased forecast lengths. HDT addresses this issue by leveraging its own trend, confirming its effectiveness in long-term predictions for high-dimensional settings.

Furthermore, to further demonstrate the effectiveness of HDT in probabilistic forecasting of high-dimensional MTS, we introduce two metrics Prediction Interval Coverage Probability (PICP) and Quantile Interval Coverage Error (QICE) from TMDM\cite{litransformer} and CARD \cite{han2022card}, which are defined as follows:
\begin{align}
\text{PICP}: &=\frac{1}{N}\sum_{n=1}^N\mathbb{I}_{y_n\geq\hat{y}_n^{\text{low}}}\cdot\mathbb{I}_{y_n\leq\hat{y}_n^{\text{high}}}, \\
\text { QICE }: &=\frac{1}{M} \sum_{m=1}^M\left|r_m-\frac{1}{M}\right|,
\end{align}
where $r_m = \frac{1}{N}\sum_{n=1}^N\mathbb{I}_{y_n\geq\hat{y}_n^{\text{low}_m}}\cdot\mathbb{I}_{y_n\leq\hat{y}_n^{\text{high}_m}}$, $\hat{y}_n^{\text{low}}$ and $\hat{y}_n^{\text{high}}$ represent the low and high percentiles, respectively. In our settting, we choose the $2.5^{th}$ and $97.5^{th}$ percentile, thus an ideal PICP value for the learned model should be~$95\%$ and we set $M=10$, and obtain the following $10$ quantile intervals (QIs) of the generated samples (generated sample $\hat{y} \in \mathcal{R}^{S \times \tau \times D}$, target $y \in \mathcal{R}^{\tau \times D}$, S is the sample size=100 of our setting): below the $10^{th}$ percentile, between the $10^{th}$ and $20^{th}$ percentiles, $\dots$, between the $80^{th}$ and $90^{th}$ percentiles, and above the $90^{th}$ percentile. From the Table~\ref{QICE}, we observe that HDT show competitive performance comparing to the TimeGrad, which demonstrate the effectiveness of HDT on the probabilistic forecasting setting.

\begin{figure*}[ht]
  \centering
  \includegraphics[width=1.0\linewidth]{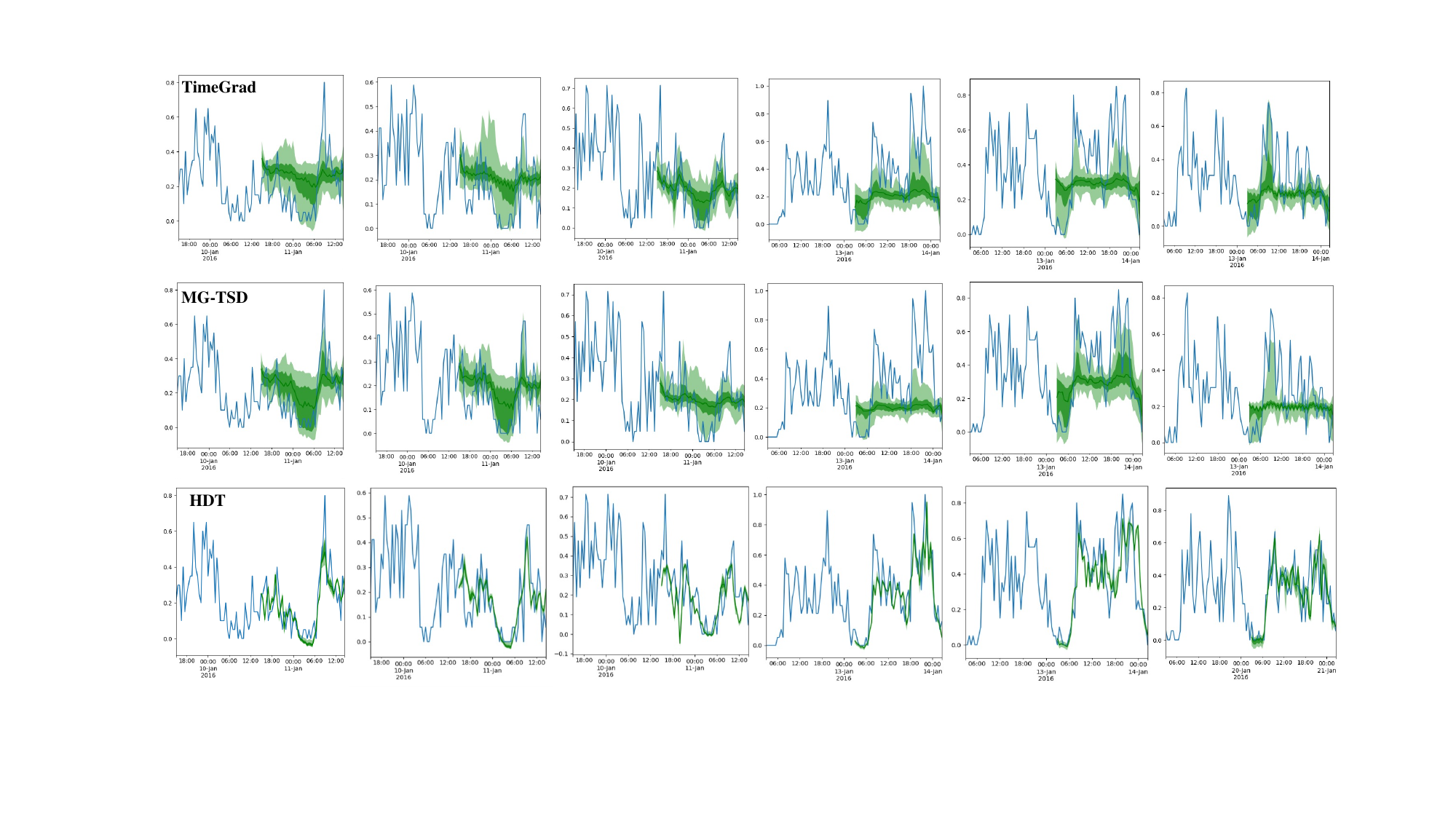}
  \vspace{-0.5in}
  \caption{Comparison of prediction intervals with TiemGrad and MG\_TSG for the Taxi dataset, which comprise 1214 dimensions. The predicted median is displayed, along with visualization of the 50\% and 90\% distribution intervals. The blue line in the graph represents the ground truth of the test sample.}
  \label{visualization_main}
\end{figure*} 

\begin{table*}[!t]
  \vskip -0.0in
  \vspace{3pt}
  \renewcommand{\arraystretch}{0.85} 
  \centering
  \resizebox{1.0\linewidth}{!}{
  \begin{threeparttable}
  \begin{small}
  \renewcommand{\multirowsetup}{\centering}
  \setlength{\tabcolsep}{1pt}
  \begin{tabular}{c|c|cc|cc|cc|cc|cc|cc|cc|cc|cc|cc|cc|cc}
    \toprule
    \multicolumn{2}{c}{\multirow{2}{*}{Models}} & 
    \multicolumn{2}{c}{\rotatebox{0}{\scalebox{0.8}{\textbf{HDT}}}} &
    \multicolumn{2}{c}{\rotatebox{0}{\scalebox{0.8}{\update{VQ-TR}}}} &
    \multicolumn{2}{c}{\rotatebox{0}{\scalebox{0.8}{MG\_TSD}}} &
    \multicolumn{2}{c}{\rotatebox{0}{\scalebox{0.8}{TSDiff}}}  &
    \multicolumn{2}{c}{\rotatebox{0}{\scalebox{0.8}{TimeDiff}}}  &
    \multicolumn{2}{c}{\rotatebox{0}{\scalebox{0.8}{SSSD}}} &
    \multicolumn{2}{c}{\rotatebox{0}{\scalebox{0.8}{{$\operatorname{D}^{3}$VAE}}}} &
    \multicolumn{2}{c}{\rotatebox{0}{\scalebox{0.8}{CSDI}}}&
    \multicolumn{2}{c}{\rotatebox{0}{\scalebox{0.8}{TimeGrad}}} &
    \multicolumn{2}{c}{\rotatebox{0}{\scalebox{0.8}{Trans-MAF}}} &
    \multicolumn{2}{c}{\rotatebox{0}{\scalebox{0.8}{DeepAR}}} 
    &\multicolumn{2}{c}{\rotatebox{0}{\scalebox{0.8}{GP-Copula}}} \\
    \multicolumn{2}{c}{} &
    \multicolumn{2}{c}{\scalebox{0.8}{\textbf{(Ours)}}} & 
    \multicolumn{2}{c}{\scalebox{0.8}{(2024)}} & 
    \multicolumn{2}{c}{\scalebox{0.8}{\citeyearpar{fan2024mg}}} & 
    \multicolumn{2}{c}{\scalebox{0.8}{\citeyearpar{kollovieh2024predict}}}  & 
    \multicolumn{2}{c}{\scalebox{0.8}{\citeyearpar{shen2023non}}}  & 
    \multicolumn{2}{c}{\scalebox{0.8}{(2023)}} & 
    \multicolumn{2}{c}{\scalebox{0.8}{\citeyearpar{li2022generative}}} & 
    \multicolumn{2}{c}{\scalebox{0.8}{\citeyearpar{tashiro2021csdi}}}& 
    \multicolumn{2}{c}{\scalebox{0.8}{\citeyearpar{rasul2021autoregressive}}} &
    \multicolumn{2}{c}{\scalebox{0.8}{\citeyearpar{rasul2020multivariate}}} &
    \multicolumn{2}{c}{\scalebox{0.8}{\citeyearpar{salinas2020deepar}}} 
    &\multicolumn{2}{c}{\scalebox{0.8}{\citeyearpar{salinas2019high}}} \\
    \cmidrule(lr){3-4} \cmidrule(lr){5-6}\cmidrule(lr){7-8} \cmidrule(lr){9-10}\cmidrule(lr){11-12}\cmidrule(lr){13-14} \cmidrule(lr){15-16} \cmidrule(lr){17-18} \cmidrule(lr){19-20} \cmidrule(lr){21-22} \cmidrule(lr){23-24} \cmidrule(lr){25-26}
    \multicolumn{2}{c}{Metric}  & \scalebox{0.78}{$\hat{\operatorname{CRPS}}$} & \scalebox{0.78}{$\hat{\text{NRMSE}}$}  & \scalebox{0.78}{$\hat{\operatorname{CRPS}}$} & \scalebox{0.78}{$\hat{\text{NRMSE}}$}  & \scalebox{0.78}{$\hat{\operatorname{CRPS}}$} & \scalebox{0.78}{$\hat{\text{NRMSE}}$}  & \scalebox{0.78}{$\hat{\operatorname{CRPS}}$} & \scalebox{0.78}{$\hat{\text{NRMSE}}$}  & \scalebox{0.78}{$\hat{\operatorname{CRPS}}$} & \scalebox{0.78}{$\hat{\text{NRMSE}}$}  & \scalebox{0.78}{$\hat{\operatorname{CRPS}}$} & \scalebox{0.78}{$\hat{\text{NRMSE}}$} & \scalebox{0.78}{$\hat{\operatorname{CRPS}}$} & \scalebox{0.78}{$\hat{\text{NRMSE}}$} & \scalebox{0.78}{$\hat{\operatorname{CRPS}}$} & \scalebox{0.78}{$\hat{\text{NRMSE}}$} & \scalebox{0.78}{$\hat{\operatorname{CRPS}}$} & \scalebox{0.78}{$\hat{\text{NRMSE}}$} & \scalebox{0.78}{$\hat{\operatorname{CRPS}}$} & \scalebox{0.78}{$\hat{\text{NRMSE}}$} & \scalebox{0.78}{$\hat{\operatorname{CRPS}}$} & \scalebox{0.78}{$\hat{\text{NRMSE}}$} & \scalebox{0.78}{$\hat{\operatorname{CRPS}}$} & \scalebox{0.78}{$\hat{\text{NRMSE}}$}\\
    \toprule
    \multirow{4}{*}{\update{\rotatebox{90}{\scalebox{0.95}{Solar}}}}
    &  \scalebox{0.78}{48} & \scalebox{0.78}{.004} & \scalebox{0.78}{.007} & \scalebox{0.78}{.005} & \scalebox{0.78}{.014} & \scalebox{0.78}{.006} & \scalebox{0.78}{.008} &     \scalebox{0.78}{.005} & \scalebox{0.78}{.007} & \scalebox{0.78}{.007} & \scalebox{0.78}{.012} & \scalebox{0.78}{.003} & \scalebox{0.78}{.009} &{\scalebox{0.78}{.004}} &{\scalebox{0.78}{.008}} &{\scalebox{0.78}{.006}} &{\scalebox{0.78}{.007}} & \scalebox{0.78}{.008} & \scalebox{0.78}{.011} &\scalebox{0.78}{.002} &\scalebox{0.78}{.009} &\scalebox{0.78}{.006} &\scalebox{0.78}{.010} &\scalebox{0.78}{.014} &\scalebox{0.78}{.007}\\ 
    & \scalebox{0.78}{96} & \scalebox{0.78}{.002} & \scalebox{0.78}{.006} & \scalebox{0.78}{.005} & \scalebox{0.78}{.009} & \scalebox{0.78}{.004} & \scalebox{0.78}{.011} & \scalebox{0.78}{.002} & \scalebox{0.78}{.011} &\scalebox{0.78}{.009} & \scalebox{0.78}{.015} &\scalebox{0.78}{.001} &\scalebox{0.78}{.005} &{\scalebox{0.78}{.004}} &{\scalebox{0.78}{.014}} & \scalebox{0.78}{.005} & \scalebox{0.78}{.009}  &\scalebox{0.78}{.004} &\scalebox{0.78}{.014} &\scalebox{0.78}{.004} &\scalebox{0.78}{.012}  &\scalebox{0.78}{.004} &\scalebox{0.78}{.007} &\scalebox{0.78}{.020} &\scalebox{0.78}{.012}\\ 
    & \scalebox{0.78}{144} & \scalebox{0.78}{.002} & \scalebox{0.78}{.005} & \scalebox{0.78}{.006} & \scalebox{0.78}{.021} & \scalebox{0.78}{.004} & \scalebox{0.78}{.016} & \scalebox{0.78}{.007}  &\scalebox{0.78}{.014} & \scalebox{0.78}{.011} & \scalebox{0.78}{.020} &\scalebox{0.78}{.002} &\scalebox{0.78}{.014}  &{\scalebox{0.78}{.003}} &{\scalebox{0.78}{.011}} & \scalebox{0.78}{.007} & \scalebox{0.78}{.011}  &\scalebox{0.78}{.011} &\scalebox{0.78}{.017} &\scalebox{0.78}{.004} &\scalebox{0.78}{.015} &\scalebox{0.78}{.008} &\scalebox{0.78}{.016} &\scalebox{0.78}{.023} &\scalebox{0.78}{.014} \\ 
    \cmidrule(lr){2-26}
    & \scalebox{0.78}{Avg} & \scalebox{0.78}{.003} & \scalebox{0.78}{.006} & \scalebox{0.78}{.005} & \scalebox{0.78}{.011} & \scalebox{0.78}{.005} & \scalebox{0.78}{.012} & \scalebox{0.78}{.005} & \scalebox{0.78}{.011} & \scalebox{0.78}{.009} & \scalebox{0.78}{.016} &\scalebox{0.78}{.002} &\scalebox{0.78}{.009}  &{\scalebox{0.78}{.004}} &{\scalebox{0.78}{.011}} & \scalebox{0.78}{.006} & \scalebox{0.78}{.009}  &\scalebox{0.78}{.011} &\scalebox{0.78}{.014} &\scalebox{0.78}{.003} &\scalebox{0.78}{.012} &\scalebox{0.78}{.006} &\scalebox{0.78}{.011}
    &\scalebox{0.78}{.019} &\scalebox{0.78}{.011} \\ 
    \midrule
    
    \multirow{4}{*}{\update{\rotatebox{90}{\scalebox{0.95}{Electricity}}}}
    &  \scalebox{0.78}{48} & \scalebox{0.78}{.002} & \scalebox{0.78}{.002}  &  \scalebox{0.78}{.004} & \scalebox{0.78}{.007} & \scalebox{0.78}{.002} & \scalebox{0.78}{.005} & \scalebox{0.78}{.001} & \scalebox{0.78}{.004} & \scalebox{0.78}{.003} & \scalebox{0.78}{.008} & \scalebox{0.78}{.003} & \scalebox{0.78}{.002} &{\scalebox{0.78}{.007}} &\scalebox{0.78}{.014} &\scalebox{0.78}{.000} &\scalebox{0.78}{.005} & \scalebox{0.78}{.002} & \scalebox{0.78}{.009} &\scalebox{0.78}{.001} &\scalebox{0.78}{.006} &{\scalebox{0.78}{.002}} &\scalebox{0.78}{.004} &\scalebox{0.78}{.004} & \scalebox{0.78}{.007}   \\ 
    & \scalebox{0.78}{96} & \scalebox{0.78}{.001} & \scalebox{0.78}{.003} & \scalebox{0.78}{.006} & \scalebox{0.78}{.011} & \scalebox{0.78}{.003} & \scalebox{0.78}{.006} & \scalebox{0.78}{.006} & \scalebox{0.78}{.012} & \scalebox{0.78}{.008} & \scalebox{0.78}{.012} &{\scalebox{0.78}{.005}} &{\scalebox{0.78}{.007}} &\scalebox{0.78}{.012} &\scalebox{0.78}{.017} & \scalebox{0.78}{.006} & \scalebox{0.78}{.005} &\scalebox{0.78}{.004} &\scalebox{0.78}{.013} &\scalebox{0.78}{.002} &\scalebox{0.78}{.006} &\scalebox{0.78}{.005} &\scalebox{0.78}{.009} &\scalebox{0.78}{.006} &\scalebox{0.78}{.014} \\ 
    & \scalebox{0.78}{144} & \scalebox{0.78}{.002} & \scalebox{0.78}{.007} & \scalebox{0.78}{.009} & \scalebox{0.78}{.006} & \scalebox{0.78}{.004} & \scalebox{0.78}{.009}  & \scalebox{0.78}{.003} & \scalebox{0.78}{.007}  & \scalebox{0.78}{.006} & \scalebox{0.78}{.019} &{\scalebox{0.78}{.007}} &{\scalebox{0.78}{.011}} &\scalebox{0.78}{.004} &\scalebox{0.78}{.021} & \scalebox{0.78}{.006} & \scalebox{0.78}{.013} &\scalebox{0.78}{.002} &\scalebox{0.78}{.014} &\scalebox{0.78}{.003} &\scalebox{0.78}{.011} &\scalebox{0.78}{.007} &\scalebox{0.78}{.008} &\scalebox{0.78} {.004} &\scalebox{0.78}{.011} \\ 
    \cmidrule(lr){2-26}
    & \scalebox{0.78}{Avg} & \scalebox{0.78}{.002} & \scalebox{0.78}{.004} & \scalebox{0.78}{.006} & \scalebox{0.78}{.008} & \scalebox{0.78}{.003} & \scalebox{0.78}{.007} & \scalebox{0.78}{.003} & \scalebox{0.78}{.008} & \scalebox{0.78}{.006} & \scalebox{0.78}{.013} &{\scalebox{0.78}{.005}} &{\scalebox{0.78}{.007}} &\scalebox{0.78}{.008} &\scalebox{0.78}{.017} & \scalebox{0.78}{.004} & \scalebox{0.78}{.008} &\scalebox{0.78}{.003} &\scalebox{0.78}{.012} &\scalebox{0.78}{.002} &\scalebox{0.78}{.011} &\scalebox{0.78}{.005} &\scalebox{0.78}{.007}  &\scalebox{0.78}{.005} &\scalebox{0.78}{.011} \\
    \midrule
    
    \multirow{4}{*}{\rotatebox{90}{\update{\scalebox{0.95}{Traffic}}}}
    &  \scalebox{0.78}{48} & \scalebox{0.78}{.001} & \scalebox{0.78}{.004} & \scalebox{0.78}{.003} & \scalebox{0.78}{.009} & \scalebox{0.78}{.003} & \scalebox{0.78}{.006} & \scalebox{0.78}{.006} & \scalebox{0.78}{.012} & \scalebox{0.78}{.004}& \scalebox{0.78}{.009}  &\scalebox{0.78}{.003} &{\scalebox{0.78}{.005}} & \scalebox{0.78}{.006} &\scalebox{0.78}{.034}& \scalebox{0.78}{-} & \scalebox{0.78}{-} &\scalebox{0.78}{.005} &\scalebox{0.78}{.004} &\scalebox{0.78}{.001} &\scalebox{0.78}{.007} &\scalebox{0.78}{.008} &\scalebox{0.78}{.005} 
    &\scalebox{0.78}{.003} &\scalebox{0.78}{.009} \\
    
    & \scalebox{0.78}{96} & \scalebox{0.78}{.003} & \scalebox{0.78}{.005} & {\scalebox{0.78}{.006}} & \scalebox{0.78}{.007} & \scalebox{0.78}{.007} & \scalebox{0.78}{.009} & \scalebox{0.78}{.003} & \scalebox{0.78}{.010}  & \scalebox{0.78}{.004} & \scalebox{0.78}{.007} &\scalebox{0.78}{.002} &\scalebox{0.78}{.004} &{\scalebox{0.78}{.005}} &{\scalebox{0.78}{.028}} & \scalebox{0.78}{-} & \scalebox{0.78}{-} &\scalebox{0.78}{.006} &\scalebox{0.78}{.012} &\scalebox{0.78}{.004} &\scalebox{0.78}{.006} &\scalebox{0.78}{.005} &\scalebox{0.78}{.007}  &\scalebox{0.78}{.007} &\scalebox{0.78}{.010} \\
    & \scalebox{0.78}{144} & \scalebox{0.78}{.004} & \scalebox{0.78}{.007} & \scalebox{0.78}{.005} & \scalebox{0.78}{.012} & \scalebox{0.78}{.004} & \scalebox{0.78}{.011} & \scalebox{0.78}{.004} & \scalebox{0.78}{.017} & \scalebox{0.78}{.005} & \scalebox{0.78}{.021} &\scalebox{0.78}{.007} &\scalebox{0.78}{.012} &{\scalebox{0.78}{.011}} & {\scalebox{0.78}{.026}} & \scalebox{0.78}{-} & \scalebox{0.78}{-} &\scalebox{0.78}{.006} &{\scalebox{0.78}{.018}} &\scalebox{0.78}{.002} &\scalebox{0.78}{.011} &\scalebox{0.78}{.007} &\scalebox{0.78}{.016} 
    &\scalebox{0.78}{.007} &\scalebox{0.78}{.016} \\
    \cmidrule(lr){2-26}
    & \scalebox{0.78}{Avg} & \scalebox{0.78}{.003} & \scalebox{0.78}{.005} & \scalebox{0.78}{.005} & \scalebox{0.78}{.009} & \scalebox{0.78}{.005} & \scalebox{0.78}{.009} & \scalebox{0.78}{.005} & \scalebox{0.78}{.013} & \scalebox{0.78}{.004} & \scalebox{0.78}{.012} &\scalebox{0.78}{.004} &{\scalebox{0.78}{.007}} &{\scalebox{0.78}{.007}} &{\scalebox{0.78}{.029}} & \scalebox{0.78}{-} & \scalebox{0.78}{-} &\scalebox{0.78}{.006} &\scalebox{0.78}{.011} &\scalebox{0.78}{.002} &\scalebox{0.78}{.008} &\scalebox{0.78}{.006} &\scalebox{0.78}{.009}  &\scalebox{0.78}{.006} &\scalebox{0.78}{.011} \\
    \midrule

    \multirow{4}{*}{\rotatebox{90}{\scalebox{0.95}{Taxi}}}
    &  \scalebox{0.78}{48} & \scalebox{0.78}{.003} & \scalebox{0.78}{.004} & \scalebox{0.78}{.005} & \scalebox{0.78}{.011} & {\scalebox{0.78}{.006}} & \scalebox{0.78}{.012} & \scalebox{0.78}{.005} & \scalebox{0.78}{.009} &\scalebox{0.78}{.011} & \scalebox{0.78}{.014}  & {\scalebox{0.78}{.003}} & {\scalebox{0.78}{.006}} &{\scalebox{0.78}{.003}} &{\scalebox{0.78}{.024}} & \scalebox{0.78}{-} & \scalebox{0.78}{-}  &\scalebox{0.78}{.003} &\scalebox{0.78}{.009} &\scalebox{0.78}{.002} &\scalebox{0.78}{.007} &\scalebox{0.78}{.004} &\scalebox{0.78}{.011} 
    &\scalebox{0.78}{.008} &\scalebox{0.78}{.007} \\
    
    & \scalebox{0.78}{96} & \scalebox{0.78}{.002} & \scalebox{0.78}{.007} & \scalebox{0.78}{.004} & \scalebox{0.78}{.009} &\scalebox{0.78}{.004} &\scalebox{0.78}{.013} & \scalebox{0.78}{.007} & \scalebox{0.78}{.005} & \scalebox{0.78}{.012} & \scalebox{0.78}{.018} & {\scalebox{0.78}{.006}} & {\scalebox{0.78}{.009}} &\scalebox{0.78}{.004} &\scalebox{0.78}{.032} & \scalebox{0.78}{-} & \scalebox{0.78}{-} &\scalebox{0.78}{.002} & \scalebox{0.78}{.007} &{\scalebox{0.78}{.004}} &{\scalebox{0.78}{.008}} &\scalebox{0.78}{.005} &\scalebox{0.78}{.009} &\scalebox{0.78}{.009} &\scalebox{0.78}{.012} \\ 
    & \scalebox{0.78}{144} & \scalebox{0.78}{.001} & \scalebox{0.78}{.010} & \scalebox{0.78}{.008} & \scalebox{0.78}{.007} & \scalebox{0.78}{.005} & \scalebox{0.78}{.016} & \scalebox{0.78}{.004} & \scalebox{0.78}{.007} & \scalebox{0.78}{.005} & \scalebox{0.78}{.022}  & {\scalebox{0.78}{.006}} & {\scalebox{0.78}{.016}}  & \scalebox{0.78}{.004} &\scalebox{0.78}{.019} & \scalebox{0.78}{-} & \scalebox{0.78}{-} & \scalebox{0.78}{.007} &\scalebox{0.78}{.013} &\scalebox{0.78}{.005} &\scalebox{0.78}{.009} &\scalebox{0.78}{.004} &\scalebox{0.78}{.007} &{\scalebox{0.78}{.015}} &\scalebox{0.78}{.017}\\ 
    \cmidrule(lr){2-26}
    & \scalebox{0.78}{Avg} & \scalebox{0.78}{.002} & \scalebox{0.78}{.007} & \scalebox{0.78}{.006} & \scalebox{0.78}{.009} & \scalebox{0.78}{.005} & \scalebox{0.78}{.013} & \scalebox{0.78}{.006} & \scalebox{0.78}{.007} & \scalebox{0.78}{.009} & \scalebox{0.78}{.018}  &{\scalebox{0.78}{.015}} &{\scalebox{0.78}{.010}} &\scalebox{0.78}{.004} &\scalebox{0.78}{.025} & \scalebox{0.78}{-} & \scalebox{0.78}{-} &\scalebox{0.78}{{.004}} &\scalebox{0.78}{{.010}} &\scalebox{0.78}{.004} &\scalebox{0.78}{.008} &\scalebox{0.78}{.005} &\scalebox{0.78}{.009}
    &\scalebox{0.78}{.011} &\scalebox{0.78}{.012} \\
    \midrule
    
    \multirow{3}{*}{\rotatebox{90}{\scalebox{0.80}{Wikipedia}}} 
    &  \scalebox{0.78}{48} & \scalebox{0.78}{.004} & \scalebox{0.78}{.005} & \scalebox{0.78}{.004} & \scalebox{0.78}{.006} & \scalebox{0.78}{.003} & \scalebox{0.78}{.008} & \scalebox{0.78}{.003} & \scalebox{0.78}{.007} & \scalebox{0.78}{.005} & \scalebox{0.78}{.011} &\scalebox{0.78}{.005} &\scalebox{0.78}{.007} &\scalebox{0.78}{.003} 
    &\scalebox{0.78}{.043} &\scalebox{0.78}{-} 
    &\scalebox{0.78}{-} & \scalebox{0.78}{.006} & \scalebox{0.78}{.015} &\scalebox{0.78}{.004} &\scalebox{0.78}{.007} &{\scalebox{0.78}{.004}} &{\scalebox{0.78}{.006}} &\scalebox{0.78}{.004} &\scalebox{0.78}{.009}  \\ 
    & \scalebox{0.78}{96} & \scalebox{0.78}{.006} & \scalebox{0.78}{.011} & \scalebox{0.78}{.007} & \scalebox{0.78}{.013} & \scalebox{0.78}{.004} & \scalebox{0.78}{.012} & \scalebox{0.78}{.004} & \scalebox{0.78}{.009} & \scalebox{0.78}{.008} & \scalebox{0.78}{.007} &{\scalebox{0.78}{.007}} &\scalebox{0.78}{.011} &\scalebox{0.78}{.006} &{\scalebox{0.78}{.051}} & \scalebox{0.78}{-} & \scalebox{0.78}{-} &\scalebox{0.78}{.007} &\scalebox{0.78}{.008} &\scalebox{0.78}{.005} &\scalebox{0.78}{.006} &\scalebox{0.78}{.008} &\scalebox{0.78}{.012} 
    &\scalebox{0.78}{.009} &\scalebox{0.78}{.011} \\
    \cmidrule(lr){2-26}
    & \scalebox{0.78}{Avg} & \scalebox{0.78}{.005} & \scalebox{0.78}{.008} & \scalebox{0.78}{.006} & \scalebox{0.78}{.010} & \scalebox{0.78}{.004} & \scalebox{0.78}{.010} & \scalebox{0.78}{.004} & \scalebox{0.78}{.008} & \scalebox{0.78}{.006} & \scalebox{0.78}{.009} &\scalebox{0.78}{.006} &\scalebox{0.78}{.009} &\scalebox{0.78}{.005} &\scalebox{0.78}{.047} & \scalebox{0.78}{-} & \scalebox{0.78}{-} &\scalebox{0.78}{.006} &\scalebox{0.78}{.011} &{\scalebox{0.78}{.005}} &{\scalebox{0.78}{.006}} &\scalebox{0.78}{.006} &\scalebox{0.78}{.009} &\scalebox{0.78}{.006} &\scalebox{0.78}{.010} \\
    \bottomrule
  \end{tabular}
    \end{small}
  \end{threeparttable}
}
  \caption{Model performance variances  on the test set $\hat{\operatorname{CRPS}}$:$\operatorname{CRPS}_{\operatorname{sum}}$, $\hat{\text{NRMSE}}$:$\text{NRMSE}_{sum}$ show baselines and our HDT model. – marks out-of-memory failures. Trans-MAF stands for Transformer-MAF.}\label{full——variance}
\end{table*}
\section{F. Details of baselines}
\label{baselines}
In our experiments, we compared HDT against 5 types of models, which are shown as follows.\\
1. Gaussian process based model
\begin{itemize}
\item GP-Copula \cite{salinas2019high}: It employs a separate LSTM unrolling for each time series, and models the joint emission distribution using a Gaussian copula with a low-rank plus diagonal covariance structure.
\end{itemize}
2. Probabilistic Deep Learning model
\begin{itemize}
\item DeepAR
\cite{salinas2020deepar}:A probabilistic model based on RNNs that learns the distribution parameters for predicting the next time point.
\end{itemize}
3. Normalizing flow based models
\begin{itemize}
\item Transformer-MAF \cite{rasul2020multivariate}: Replace the LSTM of the LSTM-MAF with the Transformer.
\end{itemize}
4. Diffusion based models
\begin{itemize}
\item TimeGrad \cite{rasul2021autoregressive}: An auto-regressive model based on the diffusion model, which is used for generating each timestamp value autoregressively. 
\item CSDI \cite{rasul2020multivariate}: A two types of Transformer based non-autoregressive diffusion model for generating multivariate time series.
\item $\operatorname{D}^3$VAE \cite{li2022generative}: A coupled diffusion probabilistic model with bidirectional variational auto-encoder (BVAE) for time series generation.
\item SSSD \cite{alcaraz2022diffusion}: Replaces the transformers in CSDI by a structured state space model to avoid the quadratic complexity issue with non-autoregressive way.
\item TimeDiff \cite{shen2023non}: A non-autoregressive diffusion model with future mixup and autoregressive initialization strategies for multivariate time series forecasting. 
\item TSDiff \cite{kollovieh2023predict}: An unconditional diffusion model with self-guidance strategy for probabilistic time series forecasting
\end{itemize}
5. Discrete vector quantization models
\begin{itemize}
\item VQ-TR \cite{rasul2023vq}: Map large sequences to a discrete set of latent representations as part of the Attention module for time series forecasting.
\end{itemize}

\section{G. Limitations and Visualizations}
\begin{figure*}[p!]
\begin{center}
   \includegraphics[width=1.0\textwidth]{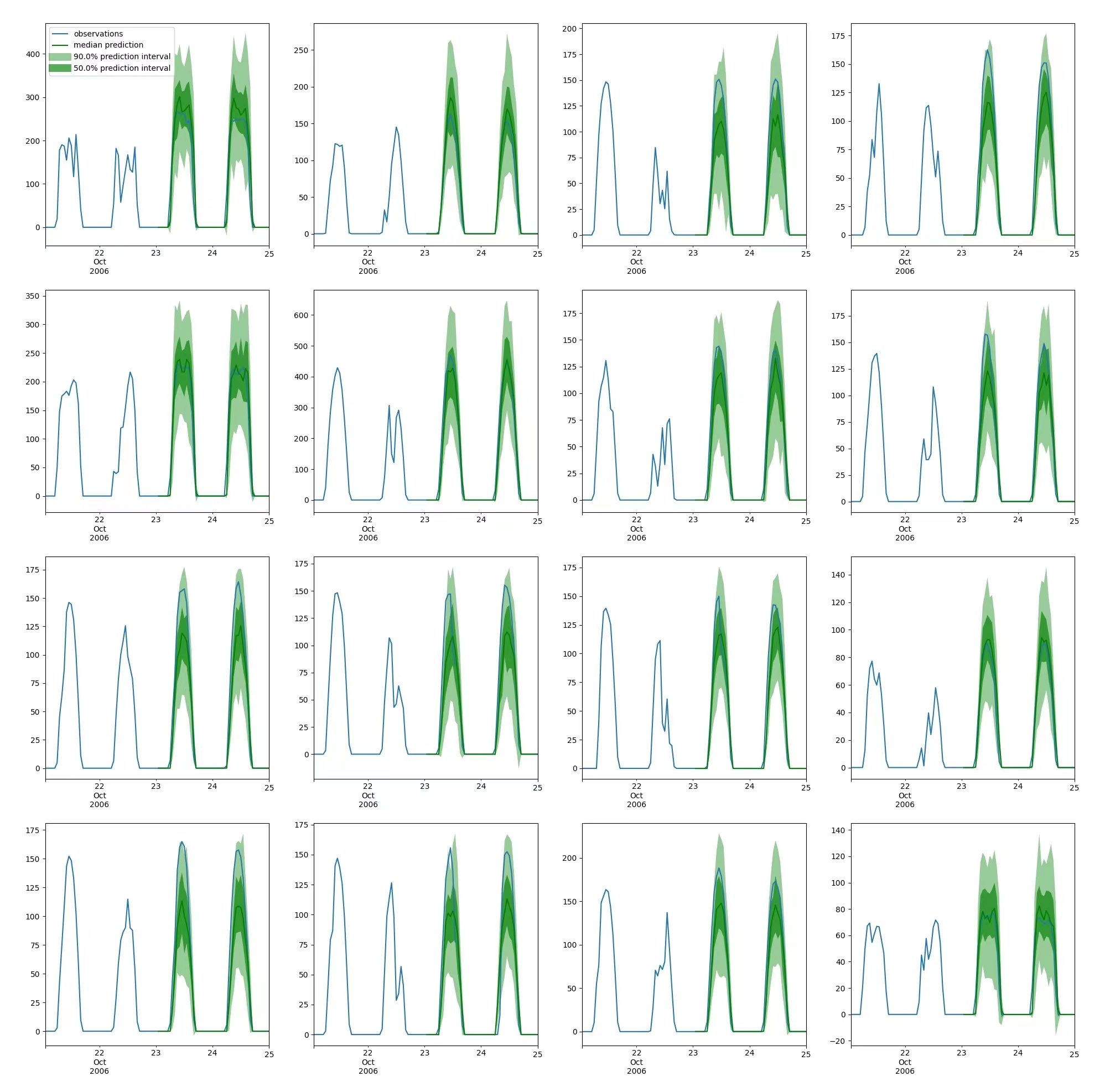}
  \vspace{-0.1in}
  \caption{The forecasting results of 16 samples from the Solar dataset with input-96-predict-48.}
  \label{1_appendix}
\end{center}
\end{figure*}

\begin{figure*}[p!]
\begin{center}
   \includegraphics[width=1.0\textwidth]{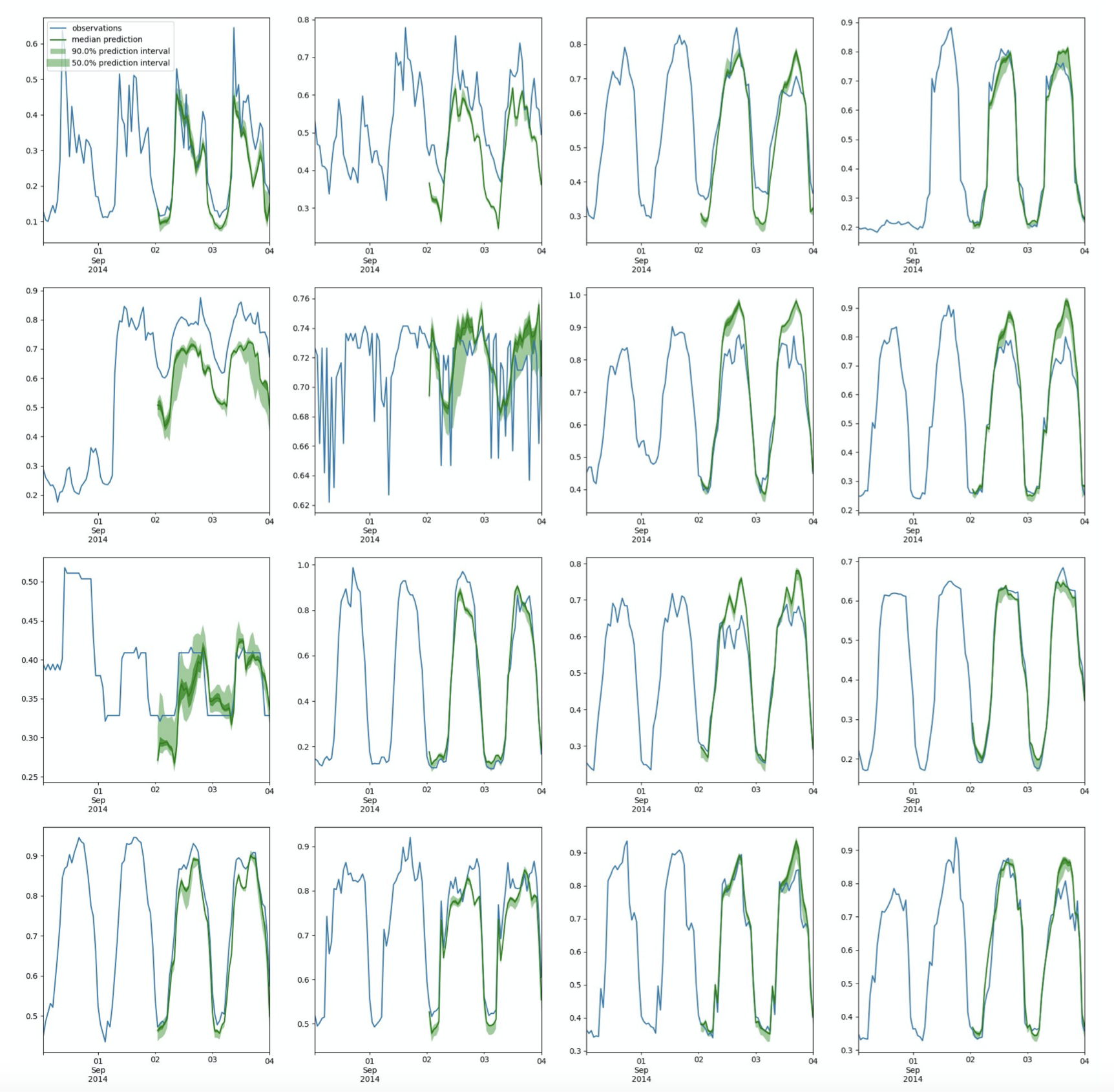}
  \vspace{-0.1in}
  \caption{The forecasting results of 16 samples from the Electricity dataset with input-96-predict-48.}
  \label{2_appendix}
\end{center}
\end{figure*}

\begin{figure*}[hp!]
\begin{center}
   \includegraphics[width=1.0\textwidth]{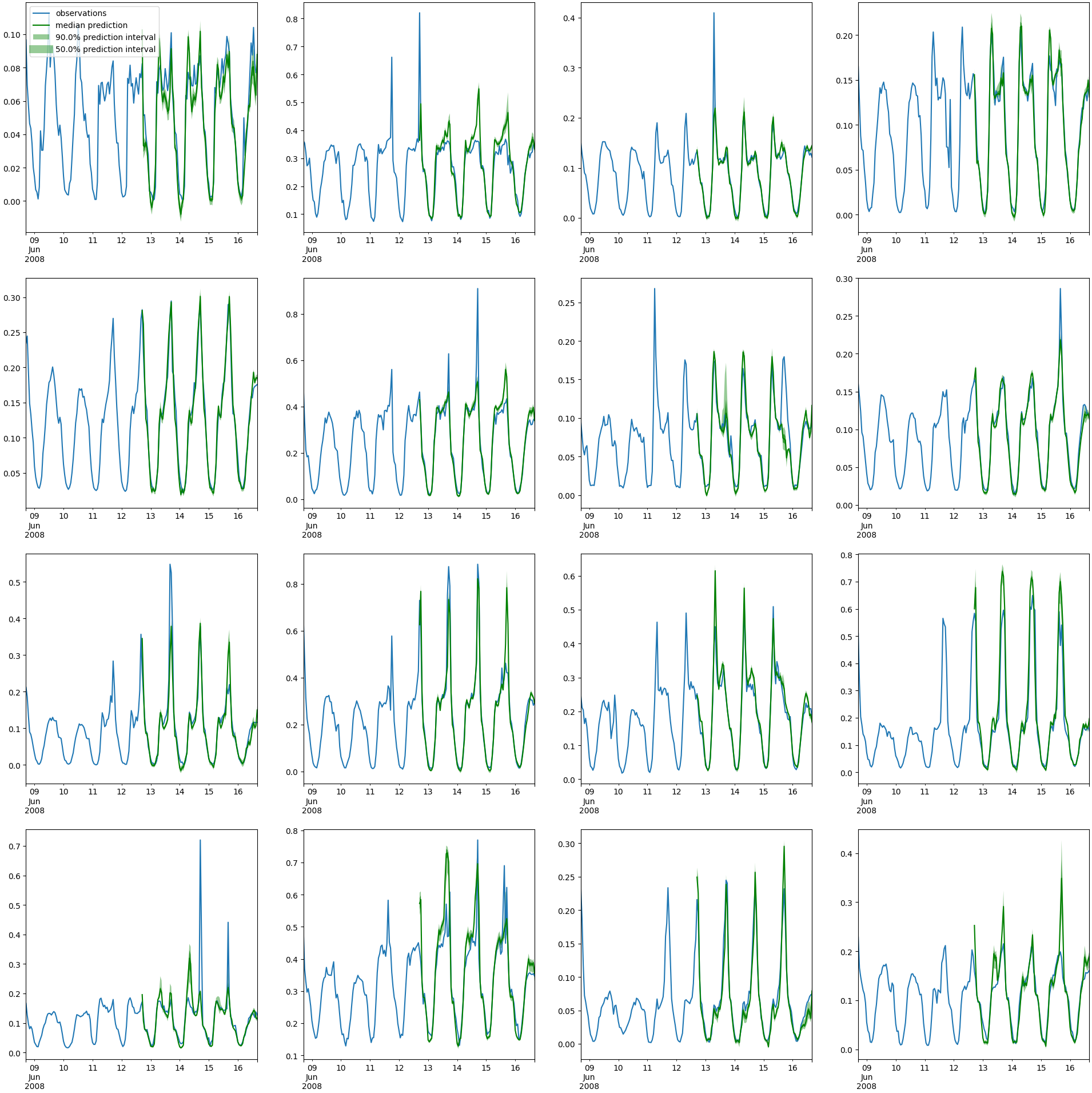}
  \vspace{-0.1in}
  \caption{The forecasting results of 16 samples from the Traffic dataset with input-96-predict-96.}
  \label{5_appendix}
\end{center}
\end{figure*}

\begin{figure*}[hp!]
\begin{center}
   \includegraphics[width=1.0\textwidth]{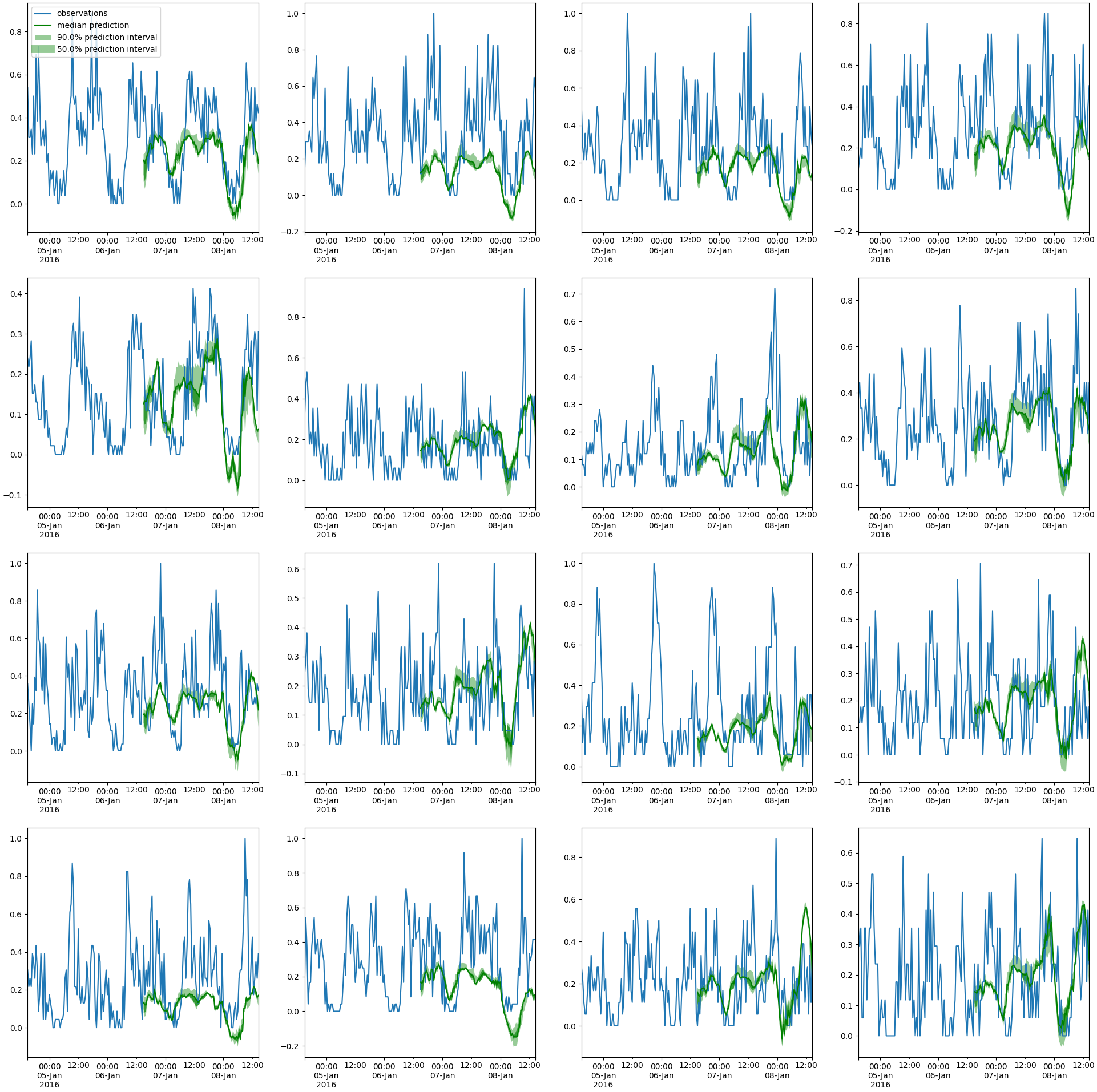}
  \vspace{-0.1in}
  \caption{The forecasting results of 16 samples from the Taxi dataset with input-96-predict-96.}
  \label{7_appendix}
\end{center}
\end{figure*}

\begin{figure*}[hp!]
\begin{center}
   \includegraphics[width=1.0\textwidth]{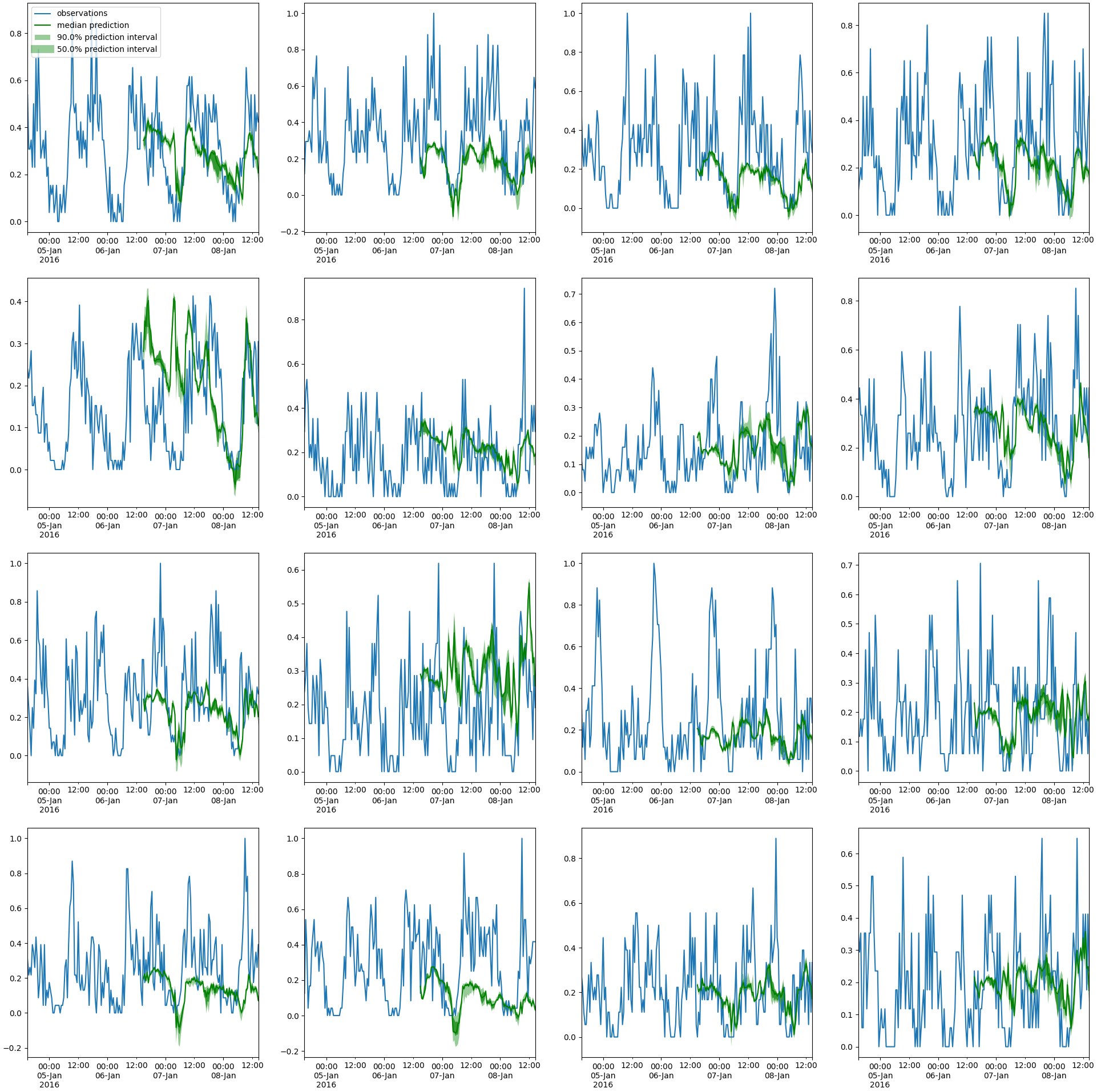}
  \vspace{-0.1in}
  \caption{The forecasting results of 16 samples from the Wikipedia dataset with input-96-predict-96.}
  \label{9_appendix}
\end{center}
\end{figure*}

\begin{figure*}[hp!]
\begin{center}
   \includegraphics[width=1.0\textwidth]{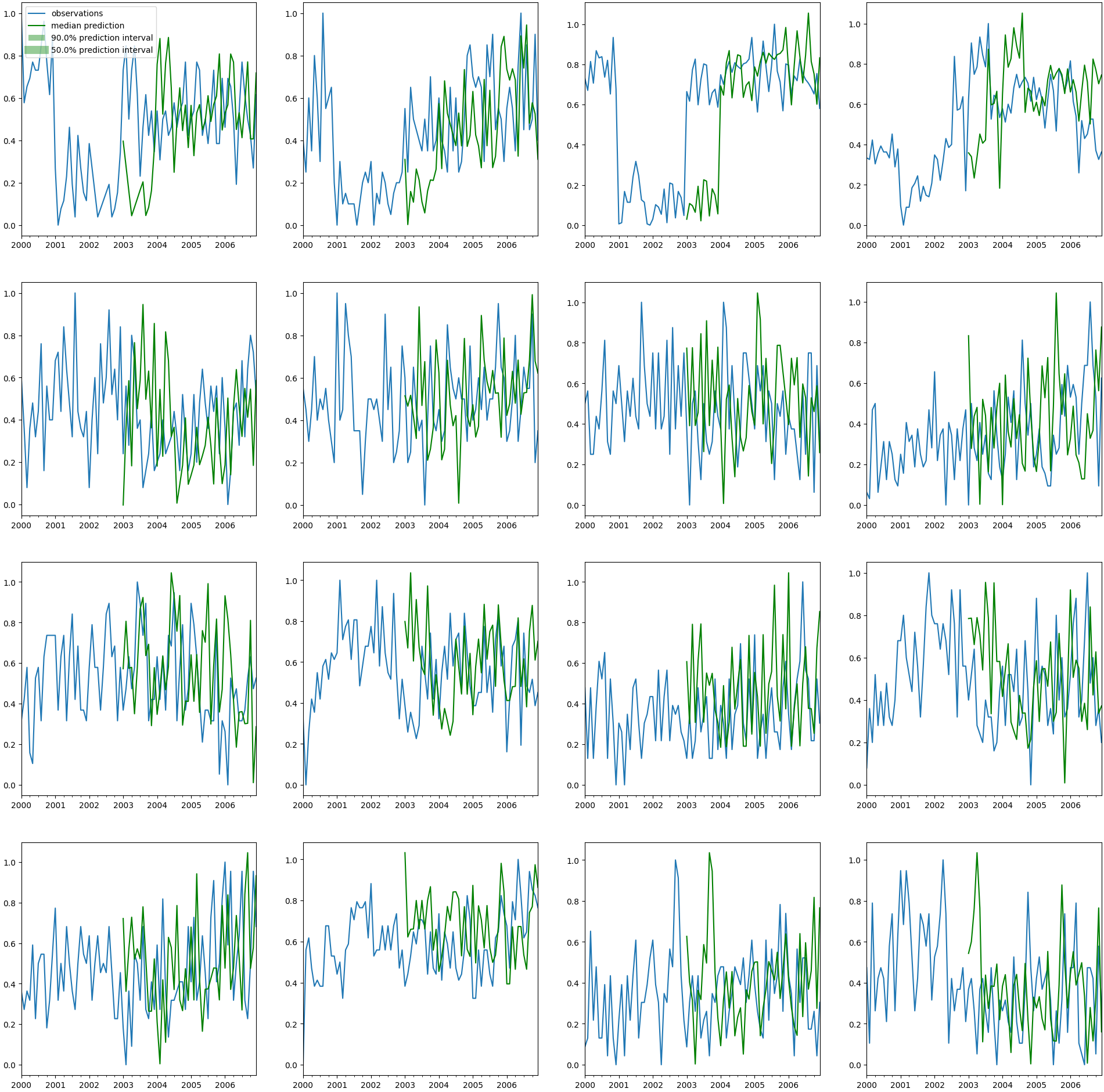}
  \vspace{-0.1in}
  \caption{The forecasting results of 16 samples from the hospital dataset with input-24-predict-48.}
  \label{10_appendix}
\end{center}
\end{figure*}

In the section, we summarize the limitation of this work and showcase ground-truths and generations on the five datasets of our main experiment, as shown in Fig.~\ref{1_appendix} to Fig.~\ref{10_appendix}. 

\noindent \textbf{Limitations:} In HDT, we need to train a separate codebook for each dataset, as we have not yet achieved the discretization of all datasets under a unified codebook setup. It is important to note that due to significant distribution differences between various time series, discretizing all datasets with a single codebook is highly challenging. In the future, we plan to draw on the approach of MOIRAI \cite{woo2024unified} to construct a unified discretized representation based on the unified time series modeling approach. 

\noindent \textbf{Visualization:} We showcase the visualized results cross five datasets with the corresponding prediction length used in our main experiment, which are shown in Fig.~\ref{1_appendix} to Fig.~\ref{10_appendix}.

\clearpage

\end{document}